\documentclass{article}

\PassOptionsToPackage{numbers, compress}{natbib}
\usepackage[preprint]{neurips_2026}

\usepackage[utf8]{inputenc} 
\usepackage[T1]{fontenc}    
\usepackage{hyperref}       
\usepackage{url}            
\usepackage{booktabs}       
\usepackage{amsfonts}       
\usepackage{nicefrac}       
\usepackage{microtype}      
\usepackage{xcolor}         
\usepackage{graphicx}
\usepackage{enumitem}
\usepackage{amsmath} 
\usepackage{float}
\usepackage{multirow}
\usepackage[table]{xcolor}
\usepackage{wrapfig}
\usepackage{pifont}
\title{EgoHMP: Achieving Precise \underline{H}uman \underline{M}otion \underline{P}rediction in 3D Scenes via \underline{Ego}centric Cues}

\author{%
  Xin Zhao$^{1}$, Pengzhan Zhou$^{1}$\thanks{The corresponding author.}, Ziyi Li$^{1}$, Riheng Jia$^{2}$, Zhida Qin$^{3}$, Yu Liu$^{4}$\\
  $^{1}$College Of Computer Science, Chongqing University\\
  $^{2}$School of Computer Science and Technology, Zhejiang Normal University\\
  $^{3}$School of Computer Science and Technology, Beijing Institute of Technology\\
  $^{4}$Department of Computing, Hong Kong Polytechnic University\\
  \texttt{xzhao@stu.cqu.edu.cn,\ pzzhou@cqu.edu.cn,\ shanksrlrobot@stu.cqu.edu.cn}\\
  \texttt{rihengjia@zjnu.edu.cn,\ zanderqin@bit.edu.cn,\ yu-y.liu@polyu.edu.hk}
}

\begin{document}
\maketitle
\vspace{-10pt}
\begin{figure}[H]
\centering
\includegraphics[width=\linewidth]{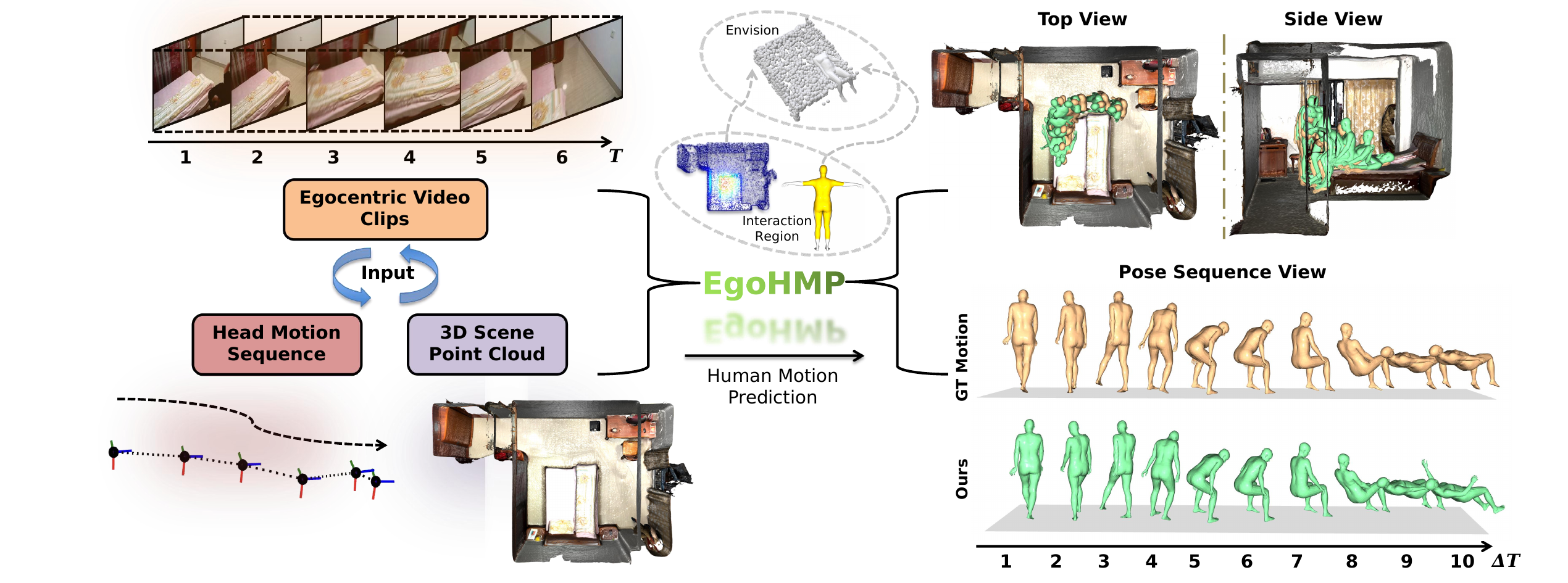}
\vspace{-15pt}
  \caption{EgoHMP captures human intent through synergistic egocentric observations and head motions, and reveals interaction concepts by combining the scene and human interaction regions, thereby envisioning the interaction, which enables precise human motion prediction in 3D scenes.}
\label{fig:teaser}
\end{figure}

\begin{abstract}
The complex interactive dependencies between human behavior and 3D scenes make accurately predicting human intent a major challenge for human motion prediction (HMP). While recent studies use human gaze to infer intent, gaze signals suffer from ambiguity, drift, and a reliance on eye-tracking hardware. Research in cognitive science indicates that egocentric views inherently encode potential interactions, while head motion serves as a crucial precursor in motion planning. Motivated by this, we propose EgoHMP, a novel framework that uses egocentric views and head motions as robust carriers of interaction intent. By translating this intent into an envisioning of future interactions, EgoHMP achieves precise HMP in 3D scenes. EgoHMP first employs a vision-motion alignment encoder to align the visual and motion feature spaces. Based on these aligned features, an egocentric intent predictor with a modality-aware modulation mechanism adaptively suppresses invalid interaction cues to predict an accurate interaction probability map. Furthermore, to enhance the stability and realism of the generated motions, we propose a contact-aware motion decoder. Specifically, inspired by the human cognitive process of planning trajectories prior to action execution, we first predict a global trajectory to serve as a self-prompt, effectively mitigating the optimization conflict in diffusion models. Subsequently, we introduce a multi-stage denoiser, which relieves motion artifacts by incorporating human contact modeling through a multi-stage optimization strategy. Finally, a GCN-based motion decoder is employed to synthesize physically plausible and semantically consistent motions. Extensive experiments demonstrate that EgoHMP achieves state-of-the-art performance. Our code is available at \url{https://github.com/zhaoxinv587/EgoHMP}.
\end{abstract}

\section{Introduction}
\label{Introduction}
Human motion prediction (HMP) aims to infer future motion sequences based on historical observations \citep{Martinez2017,Cui2020,Li2020,Pavllo2020}. It holds significant value in applications such as augmented and virtual reality (AR/VR) \citep{Zheng2022,Diomataris2024,yu2025vision}, human-computer interaction\citep{hassan2021stochastic,Mao2022}, and embodied AI \citep{Liang2023,Wang2024b,Delitzas2024,Li2024}. Unlike motion prediction in unconstrained spaces\citep{patel2025uniegomotion,li2023ego,yi2025estimating}, realistic human behaviors in 3D scenes are not isolated motion sequences; rather, they are highly intent-driven and tightly coupled with the environment. Consequently, to accurately predict motions, a model must reliably infer the underlying interaction intent. However, precisely capturing this interactive intent remains a formidable challenge.

To tackle this challenge, recent studies \citep{Zheng2022,yu2025vision,Lou2024} have attempted to use human gaze as a proxy for intent. While intuitively appealing, this approach relies on the strict premise that individuals look exclusively at their intended targets prior to interaction. In reality, the inherent drift and ambiguity of gaze signals frequently mislead models. Because distinct gaze points can correspond to entirely different motion directions and interaction targets, this ambiguity easily results in unreliable motion predictions\citep{yu2025vision}. Furthermore, acquiring high-quality 3D gaze data necessitates specialized eye-tracking devices. Consequently, these critical limitations severely restrict the predictive performance of such methods and impede their deployment in real-world scenarios.

Research in cognitive science shows that humans synthesize egocentric behaviors by coordinating the visual cortex, cerebellum, and cerebrum. This coordination intrinsically bridges visual observations, ego-motion, and conceptual understanding, thereby connecting the physical body to its surrounding environment and unveiling latent interaction intents \citep{buckner2013cerebellum,pelz2001coordination,fassihi2017transformation}. Concurrently, previous studies demonstrate that head motion serves as a crucial kinematic precursor in motor planning, consistently anticipating and guiding subsequent full-body execution \citep{patel2025uniegomotion,adolphs2003cognitive,yang2024egochoir}. Motivated by these insights, we argue that the egocentric view, coupled with its corresponding head motion sequence, acts as a more robust and accessible medium for human interaction intent than gaze signals. Specifically, the egocentric view provides rich semantic context regarding the user's focus, while head motion supplies the essential 3D spatial cues to anchor these 2D semantics into the physical world \citep{patel2025uniegomotion,yang2024egochoir,yang2024lemon}.

To instantiate this insight, we propose EgoHMP, a novel framework that achieves precise human motion prediction in 3D scenes using egocentric videos, head motion sequences, and 3D scene point clouds (Figure \ref{fig:teaser}). Specifically, EgoHMP first employs a vision-motion alignment encoder to bridge the semantic gap between visual observations and head motions. To enable the model to extract effective interaction contexts from diverse interaction cues, we propose a modality-aware modulation mechanism. This mechanism manipulates the backpropagated gradient flow, enabling the network to adaptively suppress invalid interaction cues and distill a reliable global intent token. Subsequently, an egocentric intent predictor uses scene point features to query this intent representation, predicting an accurate interaction probability map. To enhance the stability and realism of the generated motions, we propose a contact-aware motion decoder. Specifically, guided by the interaction probability map and inspired by the human cognitive process of planning paths prior to action execution, we first infer a global trajectory to serve as a self-prompt. By anchoring the diffusion process with this trajectory, the model is freed to focus on fine-grained pose synthesis. Subsequently, conditioned on multi-modal representations and local scene contexts, we introduce a multi-stage denoiser. Inspired by the insight that human contact learning is the key to bridging perception and action for envisioning interactive poses \citep{yang2024egochoir}, this denoiser progressively optimizes the generated poses by explicitly learning human contact regions, thereby generating semantically consistent and physically plausible motions.

In summary, our main contributions can be summarized as follows:
\begin{itemize}[leftmargin=*]
    \item We propose a new perspective for human motion prediction in 3D scenes, demonstrating that integrating egocentric views with head motion sequences serves as a robust proxy for inferring human interaction intent, without relying on gaze information.
    \item We introduce EgoHMP, a novel framework comprising three core components: a vision-motion alignment encoder to align 2D vision and 3D head motion features, an egocentric intent predictor to precisely ground interaction intents, and a contact-aware motion decoder to synthesize context-consistent and physically plausible motions.
    \item Extensive experiments on the GIMO \citep{Zheng2022}, GTA-IM \citep{cao2020long}, and Ego-Exo4D \citep{grauman2024ego} benchmarks demonstrate that EgoHMP achieves state-of-the-art performance in both pose and trajectory prediction accuracy, effectively alleviating geometric artifacts such as floating and penetration.
\end{itemize}

\section{Related Work}
\textbf{Human Motion Prediction in 3D Scenes.} Recent methods incorporate elements such as 3D scene geometry features, human gaze information, and object affordance into the prediction framework, enhancing prediction accuracy by integrating external context \citep{Zheng2022,yu2025vision,hassan2021stochastic,Lou2024,Han2024,xing2024scene}. SIF3D \citep{Lou2024} uses human gaze as a motion intention and models the motion-constrained space using salient point clouds derived from 3D scenes. However, it overlooks humans' subjective intention to interact with the environment, leading to predictions that lack targeted interaction behaviors and fail to accurately infer human-object or human-computer interaction actions. Subsequent works \citep{yu2025vision,Wang2024b} further incorporate object affordance, inferring interaction gestures by leveraging objects' functional attributes in conjunction with human motion history. GAP3DS \citep{yu2025vision} constructs an object-action mapping using predefined affordance labels to enhance the semantic rationality of predicted interactions. However, the above methods neglect the inherent drift and ambiguity of gaze information, rendering them prone to trajectory-prediction biases and misidentification of the intended interactive object. In contrast, EgoHMP circumvents the reliance on gaze signals. By coupling egocentric views with head motion sequences, we provide a more robust proxy for human interaction intent in 3D scenes.

\textbf{Egocentric Intent Understanding.} The egocentric perspective inherently captures the user's visual attention and functional focus, making it a powerful tool for intent understanding. Recently, egocentric intent understanding has achieved remarkable progress \citep{patel2025uniegomotion, yang2024egochoir,mascaro2023intention,roy2024interaction,wu2020learning,escobar2025egocast,wang2025ego4o}. However, a critical limitation of these approaches is their lack of explicit 3D scene perception. They primarily perceive intents within 2D visual spaces, forcing downstream tasks—such as motion prediction—to operate exclusively within the local camera coordinate system, leaving them completely oblivious to the global physical world. Consequently, predicting physically plausible and collision-free interactions in real-world 3D environments remains highly challenging for these methods. EgoHMP elegantly bridges this gap. By seamlessly integrating egocentric signals with 3D scene point clouds, EgoHMP successfully grounds 2D egocentric intents into 3D scenes.

\textbf{Diffusion Models for Motion Generation.} 
Recently, denoising diffusion probabilistic models \citep{yu2025vision,Wang2024b,ho2020denoising,tevet2022motionclip,Yuan2023} have emerged as the dominant paradigm for human motion generation due to their exceptional generative diversity and fidelity. However, we observe that forcing a single diffusion model to simultaneously regress global locomotion and local joint articulation introduces an optimization conflict. Furthermore, while most existing methods incorporate object affordances to improve physical plausibility, their success depends on the accuracy of scene segmentation. More importantly, they largely overlook the human-centric contact. To address these limitations, we first propose a self-prompting strategy. By predicting an intent-aligned global trajectory as prompt to pose synthesis, we relieve the optimization conflict. Second,  EgoHMP employs a denoiser driven by a multi-stage optimization strategy. It leverages egocentric vision-motion cues to model human-centric contact, utilizing these contact priors to refine motions. This effectively resolves physical artifacts, empowering EgoHMP to generate context-consistent and physically plausible motions.

\section{Method}
\label{Method}
\begin{figure}
\centering
\includegraphics[width=\linewidth,scale=1]{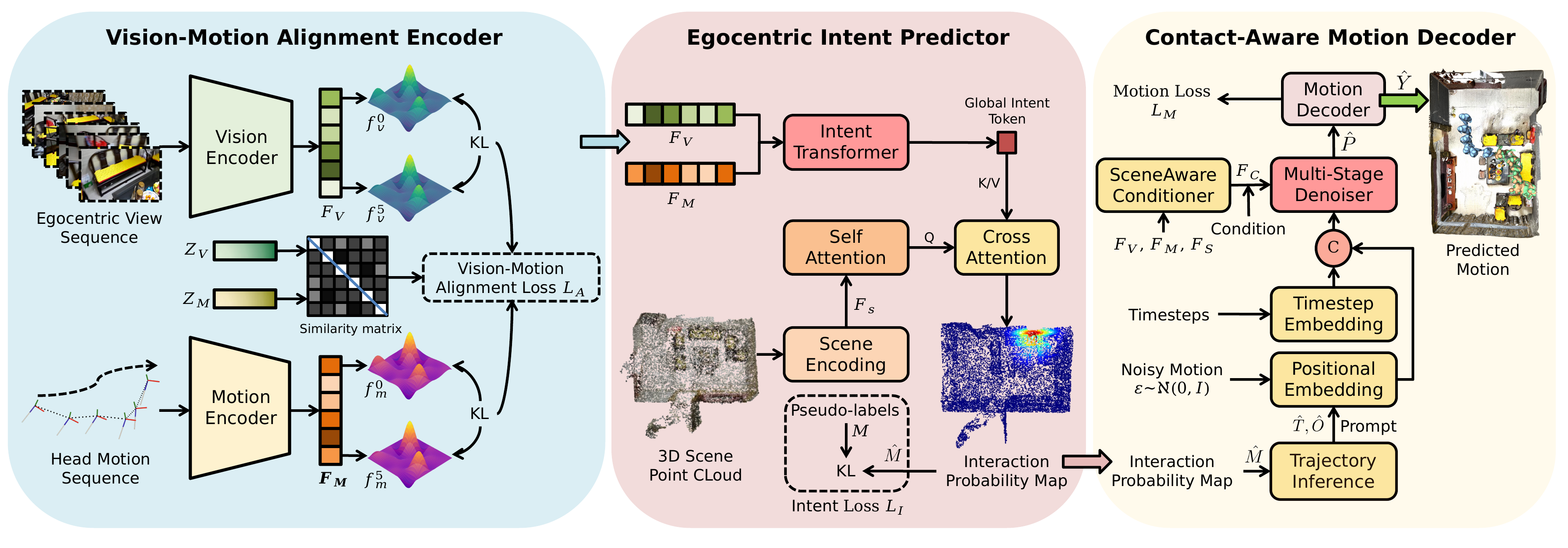}
\vspace{-15pt}
  \caption{\textbf{Pipeline of EgoHMP.} EgoHMP first employs a pre-trained vision-motion alignment encoder to extract aligned features by minimizing the discrepancy between visual and motion disparities (Sec. \ref{VMAE}). Subsequently, the egocentric intent predictor uses these aligned features to query the 3D scene point clouds, inferring an interaction probability map (Sec. \ref{EIP}). Finally, guided by this probability map, a contact-aware motion decoder synthesizes the final human motion sequence (Sec. \ref{CAMD}).}
  \label{fig:pipeline}
\end{figure}

Figure \ref{fig:pipeline} shows EgoHMP's overall architecture. EgoHMP accurately infers human intent and generates physically plausible, semantically consistent human motion predictions. 
\subsection{Problem Setup}
Given an egocentric video clip $V_{1:T} \in \mathbb{R}^{T \times H \times W \times 3}$, its corresponding head motion sequence $H_{1:T} = \{(t^h_1, o^h_1), \dots, (t^h_T, o^h_T)\}$ (where $t^h \in \mathbb{R}^3$ and $o^h \in \text{SO}(3)$ denote the head translation and orientation, respectively), and a 3D scene point cloud $S = \{p_1, \dots, p_N\} \in \mathbb{R}^{N \times 6}$ with RGB, our objective is to forecast the future motion sequence $\hat{Y}_{T+1:T+\Delta T} = \{x_{T+1}, \dots, x_{T+\Delta T}\}$. Following \citep{pavlakos2019expressive}, each pose is parameterized as $x_k = (t_k, o_k, p_k)$, comprising the translation $t_k \in \mathbb{R}^3$, the orientation $o_k \in \text{SO}(3)$, and a body pose embedding $p_k \in \mathbb{R}^{32}$. Following \citep{yu2025vision}, we define the historical observation window as $T = 6$ frames and the future prediction horizon as $\Delta T = 10$ frames to achieve long-term prediction.
Formally, we cast inference as a conditional generative process. Parameterized by $\theta$, our model $f_\theta$ synthesizes future motion by iteratively denoising a pure Gaussian noise sequence $\mathcal{Z} \sim \mathcal{N}(\mathbf{0}, \mathbf{I})$ conditioned on the multi-modal inputs:
\begin{equation}
\hat{Y}_{T+1:T+\Delta T} = \arg\max f_\theta\big(\mathcal{Z} \mid V_{1:T}, H_{1:T}, S\big).
\end{equation}

\subsection{Vision-Motion Alignment Encoder}
\label{VMAE}

\textbf{Vision Encoder.} 
The vision encoder aims to extract context-rich visual representations from egocentric video clips. Given an input sequence $V_{1:T} \in \mathbb{R}^{T \times H \times W \times 3}$, each frame is encoded using a pre-trained DINOv2 \citep{oquab2023dinov2}, followed by a linear projection to obtain frame-level embeddings $F_{V} \in \mathbb{R}^{T \times D}$, where $D$ denotes the feature dimension. To capture temporal dependencies in human motion, we introduce a learnable temporal positional encoding into $F_V$ and feed the resulting sequence into a multi-layer Transformer encoder to model correlations among frame-level features. Meanwhile, we apply average pooling over $F_V$ to obtain a sequence-level global representation, which is then projected into a lower-dimensional space $Z_{V} \in \mathbb{R}^{D_{proj}}$ via MLP layers, where $D_{proj}$ denotes the projection dimension. Finally, the encoder outputs the feature representations $\{F_{V}, Z_{V}\}$.

\textbf{Motion Encoder.} The motion encoder extracts high-dimensional motion representations from the head motion sequence $H_{1:T} \in \mathbb{R}^{T \times 12}$. We first project $H_{1:T}$ via MLP layers into high-dimensional embeddings $F_{M} \in \mathbb{R}^{T \times D}$. To maintain consistency with the vision encoder, we then inject a learnable temporal positional encoding and feed the sequence into a lightweight Transformer encoder. Since head motion signals exhibit lower complexity than vision signals, a lightweight Transformer is sufficient to capture correlations among frame-level features. Following the same design paradigm as the vision encoder, we further aggregate $F_M$ via average pooling and project it into a lower-dimensional latent space $Z_{M}$. The motion encoder thus outputs the representations $\{F_{M}, Z_{M}\}$.

\textbf{Vision-Motion Alignment Loss.} 
To align visual and motion modalities, we propose a vision-motion alignment loss $\mathcal{L}_A$ that correlates cross-modal features by restricting the local distances and global semantic alignment between their respective disparities in the feature space, allowing for mutual sensitivity to modal variations. Given $F_V, F_M$ and $Z_V, Z_M$, the objective is defined as:
\begin{equation}
\mathcal{L}_A = 
\left\| 
\mathrm{KL}\big(\mathbf{d}_M^{(1)} \,\|\, \mathbf{d}_M^{(T)}\big) - 
\mathrm{KL}\big(\mathbf{d}_V^{(1)} \,\|\, \mathbf{d}_V^{(T)}\big)
\right\|^2
+ \lambda \cdot 
\frac{1}{2} \left[
\mathrm{CE}(\mathbf{S}, \mathbf{y}) + 
\mathrm{CE}(\mathbf{S}^\top, \mathbf{y})
\right],
\end{equation}
where $\mathbf{d}^{(t)} = \mathrm{softmax}(F^{(t)} / \tau_{kl})$ is the temperature-scaled feature distribution. $\mathbf{S} = (Z_M Z_V^\top) / \tau_{ce}$ denotes the temperature-scaled pairwise cosine similarity matrix within a batch, and $\mathrm{CE}$ denotes cross-entropy guided by the matching labels $\mathbf{y}$ to attract paired vision-motion samples while repelling unmatched ones within a batch \citep{radford2021learning}. $\tau_{kl}$ and $\tau_{ce}$ are temperature scalars, and $\lambda$ balances the objectives.

\subsection{Egocentric Intent Predictor}
\label{EIP}
\textbf{Scene Encoding.} We employ PointNet++ \citep{qi2017pointnet++} as our backbone to encode the 3D scene point cloud $S \in \mathbb{R}^{N \times 6}$ into point-wise features $F_{S} \in \mathbb{R}^{N \times D_e}$ ($D_e$ denotes the feature dimension).

\textbf{Intent Transformer.} 
To synergize cross-modal features, we first project $F_V$ and $F_M$ into a shared latent space $\mathbb{R}^{T \times D_e}$. In egocentric scenarios, visual observations and head motions convey distinct intent cues with varying degrees of reliability. Indiscriminate fusion often causes reliable intent cues to be overshadowed by irrelevant noise. To address this issue and enable the network to adaptively adopt reliable cues for intent modeling, we propose a modality-aware modulation mechanism.

Specifically, we introduce learnable modulation tokens $\tau_V, \tau_M \in \mathbb{R}^{D_e}$ to scale the respective features via the Hadamard product ($\odot$), which function as a gradient interception mechanism during backpropagation. Formally, let $Z_{V}' = \tau_V \odot F_{V}$ denote the modulated visual features. 
The gradient of the total loss $\mathcal{L}$ with respect to any upstream mapping parameter $\theta$ can be derived via the chain rule, and the parameters are updated accordingly via gradient descent, as:
\begin{equation}
    \frac{\partial \mathcal{L}}{\partial \theta} = \left( \frac{\partial \mathcal{L}}{\partial Z_{V}'} \odot \tau_V \right) \frac{\partial F_V}{\partial \theta}, \theta_{t+1} \leftarrow \theta_t - \eta \left( \frac{\partial \mathcal{L}}{\partial Z_V'} \odot \tau_V \right) \frac{\partial F_V}{\partial \theta_t}
\end{equation}
where $\eta$ is the learning rate. Through this modulation, the network is able to adaptively suppress gradients from unreliable interaction cues during optimization. To finally distill a global intent, we introduce a learnable intent token $X_{intent} \in \mathbb{R}^{1 \times D_e}$ and construct a feature sequence $Z_{in} \in \mathbb{R}^{(2T+1) \times D_e}$ by incorporating temporal positional embeddings $E_{pos}$ as:
\begin{equation}
Z_{in} = \big[ X_{intent} \,\big\|\, (\tau_V \odot F_V + E_{pos}) \,\big\|\, (\tau_M \odot F_M + E_{pos}) \big],
\end{equation}
where $[\cdot \| \cdot]$ denotes concatenation. The sequence is processed by a multi-layer Transformer, and the updated intent token $X_{intent}$ is used as the global interaction representation $Z_{global} \in \mathbb{R}^{D_e}$.

\textbf{Interaction Probability Map.} 
We first use scene point features $F_{S}$ to query the processed intent sequence via cross-attention, producing fused features $F_{fused} \in \mathbb{R}^{N \times D_{e}}$. Then, we compute a semantic matching score $S_{sim} = (F_{fused} \cdot Z_{global}) / \sqrt{D_{e}}$ to measure the alignment between each point and the global intent. In parallel, a projection head predicts a point-wise interaction score $S_{mlp}$ from $F_{fused}$. Finally, the interaction probability map is defined as: $\hat{M} = S_{mlp} + S_{sim} \in \mathbb{R}^{N}$.

\textbf{Intent Loss.}
In existing benchmarks, human interaction intent lacks explicit annotations \citep{lou2024harmonizing}. To provide supervision for the intent predictor, we formulate a pseudo-label $M \in \mathbb{R}^{N}$ as follows:
\begin{equation}
    M_i = \lambda_{c} \exp\left(-\frac{\|p_i - p_{target}\|_2^2}{2\sigma_1^2}\right) + \lambda_{m} \frac{1}{\Delta T} \sum_{l=T+1}^{T+\Delta T} \left(\frac{l-T}{\Delta T}\right)^2 \exp\left(-\frac{\|p_i - t^h_l\|_2^2}{2\sigma_2^2}\right),
\end{equation}
where $p_{\text{target}}$ denotes the gt human-object contact position. The first term represents the interaction contact prior. The second term is the head motion prior, modulated by a temporal weight $(t/\Delta T)^2$ to capture the dynamic convergence of human intent. This design ensures that as time progresses toward the interaction moment, the user's head position serves as a stronger indicator of the intent. The deviations $\sigma_1$ and $\sigma_2$ are set to 0.3 and 1.0, respectively. Finally, to ensure valid probability distributions, the intent loss is defined as: $\mathcal{L}_I = \mathrm{KL}(\mathrm{softmax}(M) \,\|\, \mathrm{softmax}(\hat{M}))$.

\subsection{Contact-Aware Motion Decoder}
\label{CAMD}
\textbf{Trajectory Inference.}
To relieve optimization conflicts between local poses and global trajectories in diffusion models \citep{hassan2021stochastic,barquero2023belfusion}, we decouple them by first inferring a global trajectory (translations $\hat{T}$, rotations $\hat{O}$) as a self-prompt:
\begin{equation}
    (\hat{T},\hat{O}) = \Psi_{\text{traj}}\big([\text{Linear}(H_{1:T}) \parallel \text{Linear}(H_{mid}) \parallel \text{Linear}(\hat{p}_{target})], F_{fused} \odot W_{dist}\big),
\end{equation}
where $\Psi_{\text{traj}}$ denotes the Transformer-based decoder (detailed in the Appendix \ref{sec:motion_decoder_details}). The first term represents the coarse trajectory embeddings, constructed by concatenating the projected historical head motion $\text{Linear}(H_{1:T})$, the replicated intermediate states $\text{Linear}(H_{mid})$, and the predicted human-object contact position $\text{Linear}(\hat{p}_{target})$ ($\hat{p}_{target} \in \mathbb{R}^3$ is extracted as the point with the highest probability from $\hat{M}$). To emphasize task-relevant regions, the $F_{fused}$ are weighted by $W_{dist} \in \mathbb{R}^N$, where $W_{dist, i} = \mathcal{R}^*\big( \|p_i - \hat{p}_{target}\|_2 \big)$ using the reverse min-max normalization operator $\mathcal{R}^*$.



\textbf{Scene-Aware Conditioner.} 
Following \citep{ho2020denoising}, we formulate the pose generation as a conditional denoising diffusion process. Guided by the trajectory prompt, the diffusion model focuses primarily on the local context. Therefore, we define the condition as: $F_C = [F_V \parallel F_M \parallel F_{\text{local}}]$, where $F_{\text{local}}$ represents the local scene context. For each frame, we apply a cylindrical mask to isolate scene points along the trajectory. A small set of learnable latent queries then cross-attends to these points, efficiently distilling the point cloud into the compact local context $F_{\text{local}}$.

\textbf{Multi-Stage Denoiser.} 
During training, we use the ground-truth body poses as a clean sequence $P_0 \in \mathbb{R}^{\Delta T \times 32}$. The forward diffusion process progressively injects Gaussian noise $\epsilon \sim \mathcal{N}(\mathbf{0}, \mathbf{I})$ into $P_0$ to yield a noisy sequence $P_t$ at a random timestep $t$. 
To synthesize physically plausible motions, we propose a multi-stage denoiser $f_\phi$. The denoiser $f_\phi$ is built upon stacked Transformer blocks that extract hidden states $\mathbf{H}\in \mathbb{R}^{\Delta T \times D}$ from $P_t$, conditioned on $F_C$ and prompt $(\hat{T}, \hat{O})$. The generation process is designed into global coarse prediction stage and local contact-aware refinement stage.

In the first stage, a linear head maps $\mathbf{H}$ to a coarse pose sequence $\hat{P}_{c} \in \mathbb{R}^{\Delta T \times 32}$ to capture the fundamental motion manifold. In the second stage, recognizing that inferring specific human contact regions is crucial for enhancing egocentric motion understanding \citep{yang2024egochoir}, we introduce local contact-aware refinement. Specifically, a parallel linear head maps the embedding $\mathbf{H}$ to a human contact probability distribution $\hat{\mathcal{A}} \in [0, 1]^{\Delta T \times 10475}$, where $10475$ denotes the number of SMPL-X vertices \citep{pavlakos2019expressive}. To supervise $\hat{\mathcal{A}}$, we apply a contact loss $\mathcal{L}_{con} = \mathrm{KL}(\mathcal{A} \,\|\, \hat{\mathcal{A}})$ (detailed in Appendix). Subsequently, we project $\hat{\mathcal{A}}$ into a 22-dimensional kinematic space using SMPL-X skinning weights $\mathbf{W}$, calculating the contact importance for different joints $\hat{J} \in \mathbb{R}^{\Delta T \times 22 \times 3}$. A linear encoder $\text{Enc}$ then maps this representation into contact features $\mathbf{C} = \text{Enc}(\hat{\mathcal{A}} \mathbf{W})$. To inject $\mathbf{C}$ into the generative process without distorting the latent space, we apply Adaptive Layer Normalization (AdaLN) to obtain the modulated feature $\mathbf{H}_{mod} = \text{LayerNorm}(\mathbf{H}) \odot (1 + \gamma(\mathbf{C})) + \beta(\mathbf{C})$. Here, $\gamma(\cdot)$ and $\beta(\cdot)$ denote the learnable scale and shift parameters derived from $\mathbf{C}$. Finally, the refined pose is defined as $\hat{P}_0 = \hat{P}_{c} + \Delta P$, where the residual displacement $\Delta P$ is computed by an MLP-based refiner $f_{l}$: 
\begin{equation}
    \Delta P = f_{l}\big([\mathbf{H}_{mod} \parallel \text{stop-gradient}(\hat{P}_{c}) \parallel \hat{T} \parallel \hat{O}]\big).
\end{equation}

\textbf{Motion Decoder.} 
The final motion sequence $\hat{Y}_{T+1:T+\Delta T}$ is reconstructed via a GCN-based decoder \citep{Lou2024}. Specifically, we use the pre-trained VPoser to decode the poses $\hat{P}_0$ into joint articulations, and use the SMPL-X \citep{pavlakos2019expressive} to integrate these articulations with the trajectory $(\hat{T},\hat{O})$. The decoder then fuses these representations to a GCN model to refine the final 3D body mesh geometry.

\textbf{Motion Loss.} 
The decoder is optimized using a joint motion loss $\mathcal{L}_{M}$: $\mathcal{L}_{M} = \lambda_{t} \mathcal{L}_{traj} + \lambda_{p} \mathcal{L}_{pose} + \lambda_{c} \mathcal{L}_{con} + \lambda_{j} \mathcal{L}_{joint} + \lambda_{s} \mathcal{L}_{slide}$. Specifically, $\mathcal{L}_{traj}$ computes the L2 distance for root translations to ensure trajectory alignment. The pose loss $\mathcal{L}_{pose}$ is defined as: $\mathcal{L}_{pose} = \mathcal{L}_{coarse} + \mathcal{L}_{refine}$. $\mathcal{L}_{coarse}$ and $\mathcal{L}_{refine}$ compute the mean squared error between ground truth and coarse pose $\hat{P}_{coarse}$ as well as refined pose $\hat{P}_0$, respectively. Notably, we propose a 
joint adaptive reconstruction loss $\mathcal{L}_{joint}$:
\begin{equation}
    \mathcal{L}_{joint} = \frac{1}{\Delta T \times 22} \sum_{t=T+1}^{T+\Delta T} \sum_{j=1}^{22} \big(1 +  (\mathcal{A} \mathbf{W})_{t,j}\big) \big\| \hat{J}_{t,j} - J_{t,j} \big\|_2^2,
\end{equation}
where $\mathcal{L}_{joint}$ uses the contact importance $(\mathcal{A} \mathbf{W})_{t,j}$ to adaptively penalize joints exhibiting higher contact probabilities. Furthermore, to relieve sliding artifacts, we introduce a slide loss $\mathcal{L}_{slide}$. It penalizes the inter-frame velocity of joints conditioned on their contact importance:
\begin{equation}
    \mathcal{L}_{slide} = \frac{1}{(\Delta T - 1) \times 22} \sum_{t=T+2}^{T+\Delta T} \sum_{j=1}^{22} (\mathcal{A} \mathbf{W})_{t,j} \big\| \hat{J}_{t,j} - \hat{J}_{t-1,j} \big\|_2^2.
\end{equation}

\begin{table}[!t]
    \centering
    \caption{\textbf{Quantitative Results.} \textbf{T-P} and \textbf{T-E} denote Traj-Path and Traj-End. \textbf{M-P} and \textbf{M-E} denote MPJPE-Path and MPJPE-End. Intent localization is evaluated via \textbf{TDE@$\boldsymbol{K}$} for $K \in \{1, 50, 100\}$. \textbf{Pen.} denotes penetration error. All metrics are reported in millimeters (mm) except for \textbf{Pen.} (in $\%$). Lower values indicate better performance ($\downarrow$). The best results are highlighted in \textbf{bold}.}
    \resizebox{\textwidth}{!}{
    \begin{tabular}{l cccc ccc c cccc ccc c} 
        \toprule
        \multirow{3}{*}{\textbf{Method}} 
        & \multicolumn{8}{c}{\textbf{GIMO} \citep{Zheng2022}} 
        & \multicolumn{8}{c}{\textbf{GTA-IM} \citep{cao2020long}} \\  
        \cmidrule(lr){2-9} \cmidrule(lr){10-17}
        & \multicolumn{4}{c}{Motion Metrics $\downarrow$} 
        & \multicolumn{3}{c}{Intent Metrics $\downarrow$}
        & \multicolumn{1}{c}{Phys. $\downarrow$}
        & \multicolumn{4}{c}{Motion Metrics $\downarrow$} 
        & \multicolumn{3}{c}{Intent Metrics $\downarrow$}
        & \multicolumn{1}{c}{Phys. $\downarrow$} \\
        \cmidrule(lr){2-5} \cmidrule(lr){6-8} \cmidrule(lr){9-9} \cmidrule(lr){10-13} \cmidrule(lr){14-16} \cmidrule(lr){17-17}
        & T-P & T-E & M-P & M-E & @1 & @50 & @100 & Pen. & T-P & T-E & M-P & M-E & @1 & @50 & @100 & Pen. \\
        \midrule
        \multicolumn{17}{l}{\textit{\textbf{Gaze-based}}} \\ 
        BiFu \citep{Zheng2022}   & 661 & 727 & 167.8 & 205.0 & 415.8 & 405.3 & 398.7 & 3.73 & 683 & 1066 & 186.4 & 252.6 & 455.2 & 445.6 & 438.1 & 3.95 \\  
        SIF3D \citep{Lou2024}    & 590 & 666 & 156.6 & 195.7 & 345.2 & 336.8 & 330.4 & 4.59 & 626 &  836 & 164.9 & 227.7 & 385.6 & 378.2 & 371.9 & 4.82 \\  
        GAP3DS \citep{yu2025vision}& 575 & 623 & 141.2 & 171.4 & 210.5 & 205.1 & 201.3 & 0.95 & 610 &  799 & 155.0 & 210.1 & 235.4 & 228.6 & 224.5 & 1.12 \\  
        \midrule
        \multicolumn{17}{l}{\textit{\textbf{Non-Gaze-based}}} \\
        Auxformer \citep{Xu2023} & 688 & 893 & 173.5 & 205.4 & 620.5 & 612.3 & 605.4 & 6.65 & 688 & 1078 & 187.8 & 262.0 & 685.2 & 678.5 & 670.3 & 7.10 \\  
        MDP \citep{xing2024scene}     & 612 & 679 & 157.1 & 190.4 & 512.4 & 502.8 & 495.6 & 4.77 & 640 &  860 & 162.5 & 222.5 & 560.3 & 551.2 & 545.8 & 5.05 \\  
        \rowcolor{gray!15} 
        \textbf{EgoHMP (Full)}   &\textbf{481} &\textbf{545} &\textbf{114.9} &\textbf{138.0} &\textbf{161.3} &\textbf{153.2} &\textbf{148.6} &\textbf{0.81} &\textbf{522} &\textbf{623} &\textbf{129.6} &\textbf{177.2} &\textbf{182.5} &\textbf{176.4} &\textbf{174.0} &\textbf{0.91} \\  
        \bottomrule
    \end{tabular}
    }
    \label{tab:quantitative_comparison}
\end{table}

\section{Experiment}
\label{Experiment}
\subsection{Experimental Setup}
\textbf{Datasets: (1) GIMO} \citep{Zheng2022} has $\sim$129K frames of human pose data, including three core types: whole-body SMPL-X poses, LiDAR-scanned 3D geometric scenes (14 scenes, 10–20 objects/scene, $\sim$500K vertices/scene), and 3D gaze coordinates. Following \citep{yu2025vision,Lou2024}, we downsample sequences to 2fps and adopt the original train-test split. \textbf{(2) GTA-IM} \citep{cao2020long} has $\sim$1M frames of human pose data, with segmentation and camera pose annotations, covering 10 indoor scenes, 50 unique characters, 22 walking styles, and diverse actions. \textbf{(3) Ego-Exo4D} \citep{grauman2024ego} is a large-scale dataset containing over 1,400 hours of synchronized egocentric and exocentric videos. It captures skilled human activities across diverse real-world environments, providing rich annotations including high-fidelity 3D whole-body and hand poses, eye gaze, camera trajectories, and 3D scene point clouds.

\textbf{Metrics.} We use six metrics to evaluate EgoHMP: (1) \textbf{Traj-Path} and (2) \textbf{Traj-End} \citep{Zheng2022} measure the average and final endpoint deviations of the trajectory. (3) \textbf{MPJPE-Path} and (4) \textbf{MPJPE-End} \citep{Aksan2021a} measure the mean joint positional errors of the whole pose sequence and final pose. (5) \textbf{TDE@$\boldsymbol{K}$} evaluates intent localization accuracy via the minimum L2 distance between the ground-truth interaction point and the top-$K$ predicted interaction points. (6) \textbf{Penetration Error} calculates the percentage of body intersecting the scene via SDF \citep{park2019deepsdf}, quantifying physical plausibility.

\textbf{Baselines.} We compare EgoHMP against a comprehensive set of SoTA methods, which can be categorized into two main groups. For scene-aware methods, we compare against both non-gaze-based methods (Auxformer \citep{Xu2023}, MDP \citep{xing2024scene}) and gaze-based methods (BiFu \citep{Zheng2022}, SIF3D \citep{Lou2024}, GAP3DS \citep{yu2025vision}). To further validate EgoHMP, we benchmark against the latest egocentric-driven approaches (EgoEgo \citep{li2023ego}, UniEgoMotion \citep{patel2025uniegomotion}, EgoCast \citep{escobar2025egocast}, EgoAllo \citep{yi2025estimating}). Notably, these egocentric-driven approaches do not explicitly condition on 3D scene. Therefore, to ensure fairness comparison, we first evaluate the pose error within the local coordinate space (w/o scene-aware). Subsequently, we re-evaluate them after aligning their predicted local poses with the ground-truth global trajectories in the 3D scene (w/ scene-aware). EgoHMP is evaluated using its own predicted trajectories.

\subsection{Main Experimental Results and Analysis}
\begin{wraptable}{r}{8cm} 
    \centering
    \vspace{-19pt}
    \caption{\textbf{Quantitative Results on Ego-Exo4D} \citep{grauman2024ego}.} 
    \label{tab:kinematic_results}
    \footnotesize
    \setlength{\tabcolsep}{4pt} 
    \begin{tabular}{lccc}
    \toprule
    Method & Scene-Aware & M-P (mm) $\downarrow$ & M-E (mm) $\downarrow$ \\
    \midrule
    EgoEgo \citep{li2023ego} & \ding{55} &315.0  &401.3  \\
    \rowcolor{gray!15} 
    EgoEgo \citep{li2023ego} & \ding{51} & 263.8 &335.4  \\
    EgoCast \citep{escobar2025egocast} & \ding{55} &225.8  & 281.6 \\
    \rowcolor{gray!15} 
    EgoCast \citep{escobar2025egocast} & \ding{51} & 153.6 & 193.7 \\
    EgoAllo \citep{yi2025estimating} & \ding{55} &\textbf{207.4}  &\textbf{268.8}  \\
    \rowcolor{gray!15} 
    EgoAllo \citep{yi2025estimating} & \ding{51}  &144.7  &178.4\\
    UniEgoMotion \citep{patel2025uniegomotion} & \ding{55} & 216.4 &285.5 \\
    \rowcolor{gray!15} 
    UniEgoMotion \citep{patel2025uniegomotion} & \ding{51} &158.5  &192.5  \\
    \textbf{EgoHMP (Base)} & \ding{55} &213.5 &276.2  \\ 
    \rowcolor{gray!15} 
    \textbf{EgoHMP (Full)} & \ding{51} & \textbf{128.6} & \textbf{157.0} \\ 
    \bottomrule
    \end{tabular}
    \vspace{-10pt}
\end{wraptable}
Table \ref{tab:quantitative_comparison} presents a quantitative comparison between the proposed EgoHMP and state-of-the-art baselines on both datasets. The results demonstrate that EgoHMP outperforms all competing methods across all evaluated metrics, validating the effectiveness of our approach. Notably, compared to the best-performing gaze-based baseline, GAP3DS \citep{yu2025vision}, EgoHMP achieves consistent improvements across all metrics and secures the lowest intent localization error (TDE@$K$) on both benchmarks. While gaze-based methods generally outperform non-gaze approaches, they remain  constrained by the inherent drift and ambiguity of human gaze signals, limiting the potential for further accuracy gains. Conversely, non-gaze methods, such as MDP \citep{xing2024scene}, yield significantly higher trajectory and intent errors due to a complete lack of explicit intent localization. By substituting gaze signals with egocentric vision-motion cues, EgoHMP successfully constructs accurate interaction probability maps. This inferred intent serves as an anchor for trajectory prediction, thereby generating precise and collision-free trajectories. Subsequently, acting as a self-prompt, this trajectory alleviates the global generation burden on the diffusion model, enabling it to focus on synthesizing context-consistent and physically plausible local poses.

As shown in Table \ref{tab:kinematic_results}, we further evaluate EgoHMP on Ego-Exo4D \citep{grauman2024ego}. Under the scene-agnostic setting (\ding{55}), our model does not exhibit an absolute advantage. In this context, kinematics-focused baselines like EgoAllo \citep{yi2025estimating} achieve robust results, as our framework relies on 3D scene geometry to generate context-consistent motions. However, to demonstrate the importance of 3D scene understanding, we establish the scene-aware setting (\ding{51}). Since most baselines cannot ingest 3D scene geometry, we simulate an ideal scene constraint for them by aligning their predictions with the ground-truth trajectories. In contrast, EgoHMP uses its predicted trajectories. Although ground-truth trajectories significantly improve baseline accuracy, their pose errors remain higher than ours. This demonstrates EgoHMP effectively integrates scene context to constrain and rectify predicted poses, thereby achieving better performance.

\begin{table}[t]
    \centering
    \caption{\textbf{Quantitative Ablations.} $F_V$: egocentric view signal; $\mathbf{G}$: using the gaze signal instead of the egocentric view signal; $\tau_V,\tau_M$: modality-aware modulation; $\text{S}_{pm}$: self-prompted mechanism; $F_{\text{local}}$: scene-aware conditioner; $\mathcal{D}_{ms}$: multi-stage denoiser; $\text{D}_{m}$: GCN-based motion decoder; $\mathcal{L}_A/\mathcal{L}_{joint}/\mathcal{L}_{slide}$: loss functions. \ding{55} means without.}
    \label{tab:ablation}
    \footnotesize
    \resizebox{\linewidth}{!}{
    \begin{tabular}{lcccccccccc>{\columncolor{gray!10}}c}
        \toprule
        \textbf{Metric} $\downarrow$ & \ding{55} $F_V$  & $\mathbf{G}$ & \ding{55} $\tau_V, \tau_M$ & \ding{55} $\text{S}_{pm}$ & \ding{55} $F_{\text{local}}$ & \ding{55} $\mathcal{D}_{ms}$ & \ding{55} $\text{D}_{m}$ & \ding{55} $\mathcal{L}_{A}$ & \ding{55} $\mathcal{L}_{joint}$ & \ding{55} $\mathcal{L}_{slide}$ & \textbf{Full} \\
        \midrule
        \textbf{T-P} & 676  & 589 & 501 & 612 & 482 & 483 & 483 & 558 & 484 & 483 & \textbf{481} \\
        \textbf{T-E} & 784  & 643 & 576 & 691 & 547 & 548 & 546 & 639 & 546 & 547 & \textbf{545} \\
        \textbf{M-P} & 168.9  & 145.9 & 130.4 & 150.4 & 135.3 & 134.7 & 118.6 & 132.6 & 122.8 & 115.4 & \textbf{114.9} \\
        \textbf{M-E} & 216.8 & 171.7 & 150.6 & 182.2 & 167.2 & 165.8 & 142.0 & 161.4 & 153.5 & 140.9 & \textbf{138.0} \\
        \bottomrule
    \end{tabular}
    }
\end{table}
\begin{figure}
\centering
\includegraphics[width=\linewidth,scale=1]{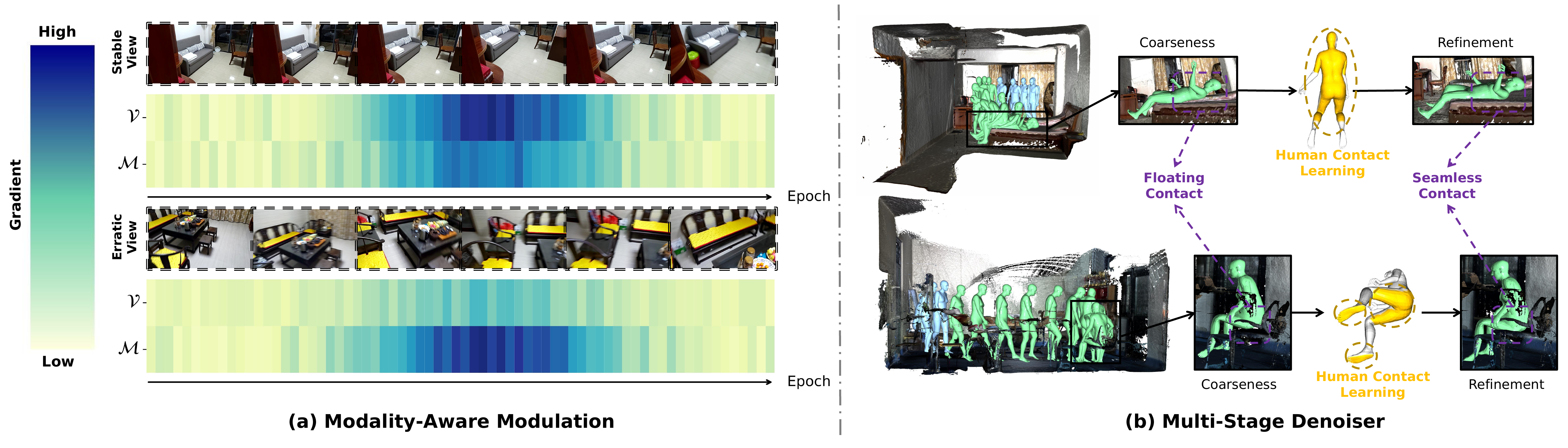}
    \vspace{-17pt}
    \caption{\textbf{Visualization Ablations.} (a) The gradient flow of the visual ($\mathcal{V}$) and motion ($\mathcal{M}$) modalities across 70 sampled training epochs.  (b) The refinement process of the multi-stage denoiser.}
  \label{fig:ablation}
\end{figure}

\subsection{Ablation Study}

We conduct extensive ablation studies on GIMO \citep{Zheng2022} and the results are summarized in Table \ref{tab:ablation}.

\textbf{Input Signals.} 
Removing the egocentric view signal (\ding{55} $F_V$) blinds the model to the scene semantics, causing a deterioration in prediction accuracy. Furthermore, replacing $F_V$ with gaze signals ($\mathbf{G}$) results in sub-optimal performance. This demonstrates that gaze signals are easily distracted by task-irrelevant objects, whereas the egocentric view provides a more robust context for intent modeling.

\textbf{Intent Modeling.} 
We first ablate the vision-motion alignment loss (\ding{55} $\mathcal{L}_{A}$). Its removal causes a significant drop in accuracy, indicating that bridging the semantic gap between visual and motion disparities is crucial for reliable intent prediction. Additionally, without the modality-aware modulation (\ding{55} $\tau_V, \tau_M$), the model fails to intercept gradients from unreliable cues, leading to degradation in prediction accuracy. As illustrated in Figure \ref{fig:ablation} (a), under a stable egocentric view, the visual observations are clean and goal-directed, generating a high gradient flow in the visual branch. This enables both modalities to update normally and maintain a healthy balance. However, under an erratic view characterized by drastic viewpoint shifts, the visual observations fail to maintain a stable focus on the target object and become cluttered with multiple irrelevant objects, rendering the interaction intent ambiguous. Regulated by the modality-aware modulation mechanism, the network dampens magnitude of the visual gradients. Consequently, the optimization focus shifts toward head motions to extract reliable interaction contexts, thereby achieving robust intent modeling.

\textbf{Motion Generation and Refinement.} 
When the self-prompted mechanism is removed (\ding{55} $\text{S}_{pm}$), the diffusion model suffers from optimization conflicts, resulting in a significant drop in motion metrics. Removing the scene-aware conditioner (\ding{55} $F_{\text{local}}$) causes the pose errors to surge sharply. This validates that the fine-grained local scene geometry is indispensable for generating physically plausible local poses. Removing the multi-stage denoiser (\ding{55} $\mathcal{D}_{ms}$) deprives the model of human contact modeling, causing noticeable pose error. As shown in Figure \ref{fig:ablation} (b), in the initial stage, the predicted poses suffer from artifacts such as floating. By incorporating human contact learning, the model leverages perceived human contact regions to correct local joint articulations, thereby generating physically plausible poses. Moreover, ablating the joint adaptive reconstruction loss (\ding{55} $\mathcal{L}_{joint}$) allows the network to bypass penalties on highly-probable contact regions. Similarly, without the slide loss (\ding{55} $\mathcal{L}_{slide}$) or the GCN-based decoder (\ding{55} $\text{D}_{m}$), the model loses micro-level geometric refinements, which are essential for relieving slide and penetration.
\begin{table}[t]
\centering
\caption{\textbf{Robustness Evaluation.} (1) Foot Sliding (FS) penalizes abnormal horizontal shifts of ankle and toe vertices when they fall below contact height thresholds (80mm and 40mm). (2) Ground Contact Deviation (GCD) captures floating/penetration artifacts via the absolute minimum vertical distance between supporting vertices and the local scene surface. (3) Human Contact Distance (HCD) evaluates interaction quality by measuring the average Euclidean distance from the intended contact vertices to the intended scene surface.}
\label{tab:robustness}
    \footnotesize
\resizebox{\textwidth}{!}{
\begin{tabular}{llcccccccc}
\toprule
\textbf{Setting} & \textbf{Perturbation}
& T-P $\downarrow$
& T-E $\downarrow$
& M-P $\downarrow$
& M-E $\downarrow$
& Pen. $\downarrow$
& FS $\downarrow$
& GCD $\downarrow$
& HCD $\downarrow$\\
\midrule
    \rowcolor{gray!15} 
Clean Input
& Original Data
& \textbf{481}
& \textbf{545}
& \textbf{114.9}
& \textbf{138.0}
& \textbf{0.81}
& \textbf{3.70}
& \textbf{36.4}
& \textbf{43.5} \\
\midrule

\multirow{2}{*}{Head-Pose Noise}
& Gaussian, $\sigma = 0.05$
& 487 & 553 & 120.2 & 142.1
& \textbf{0.81} & 3.71 & 36.6 & 44.9 \\

& Gaussian, $\sigma = 0.10$
& 505 & 572 & 128.6 & 155.7
& 0.83 & 3.74 & 37.0 & 47.3 \\
\midrule

\multirow{2}{*}{Calibration Error}
& Offset (5 cm, $2^\circ$)
& 511 & 579 & 131.8 & 161.5
& 0.83 & 3.72 & 36.9 & 51.5 \\

& Offset (10 cm, $5^\circ$)
& 524 & 594 & 139.3 & 170.4
& 0.88 & 3.76 & 37.5 & 54.0 \\
\midrule

\multirow{2}{*}{Incomplete Geo.}
& 30\% Points Dropped
& 490 & 565 & 119.3 & 144.2
& 0.86 & 3.71 & 36.5 & 44.7 \\

& 50\% Points Dropped
& 533 & 621 & 145.4 & 174.0
& 0.91 & 3.74 & 36.8 & 56.2 \\

\bottomrule
\end{tabular}
}
\end{table}

\subsection{Robustness Evaluation}
To further evaluate the robustness of EgoHMP, we simulate three types of perturbations: (1) Head-Pose Noise, where Gaussian noise is injected into head motions; (2) Calibration Error, where spatial offsets are applied  to simulate sensor misalignment; and (3) Incomplete Geometry, where the scene points are randomly removed to simulate incomplete scene scans.

As shown in Table~\ref{tab:robustness}, EgoHMP maintains relatively stable performance across  different perturbation levels, demonstrating its robustness to noisy head poses, calibration errors, and incomplete  scene geometry. This robustness benefits from the modality-aware modulation, which dynamically suppresses unreliable modalities by adjusting gradient flows, allowing complementary cues to maintain stable intent inference. Furthermore, the intent predictor and contact decoder are supervised via Gaussian-smoothed continuous probability maps rather than hard labels, preventing the network from overfitting to granular 3D details and enabling it to generalize well to sparse or lower-quality point clouds. The scene-aware conditioner further improves robustness by extracting local scene features within a bounded spatial region around the predicted trajectory, making the model less sensitive to occlusions and missing regions outside the interaction area.

\subsection{Visualization Analysis}
\begin{figure}
\centering
\includegraphics[width=\linewidth,scale=1]{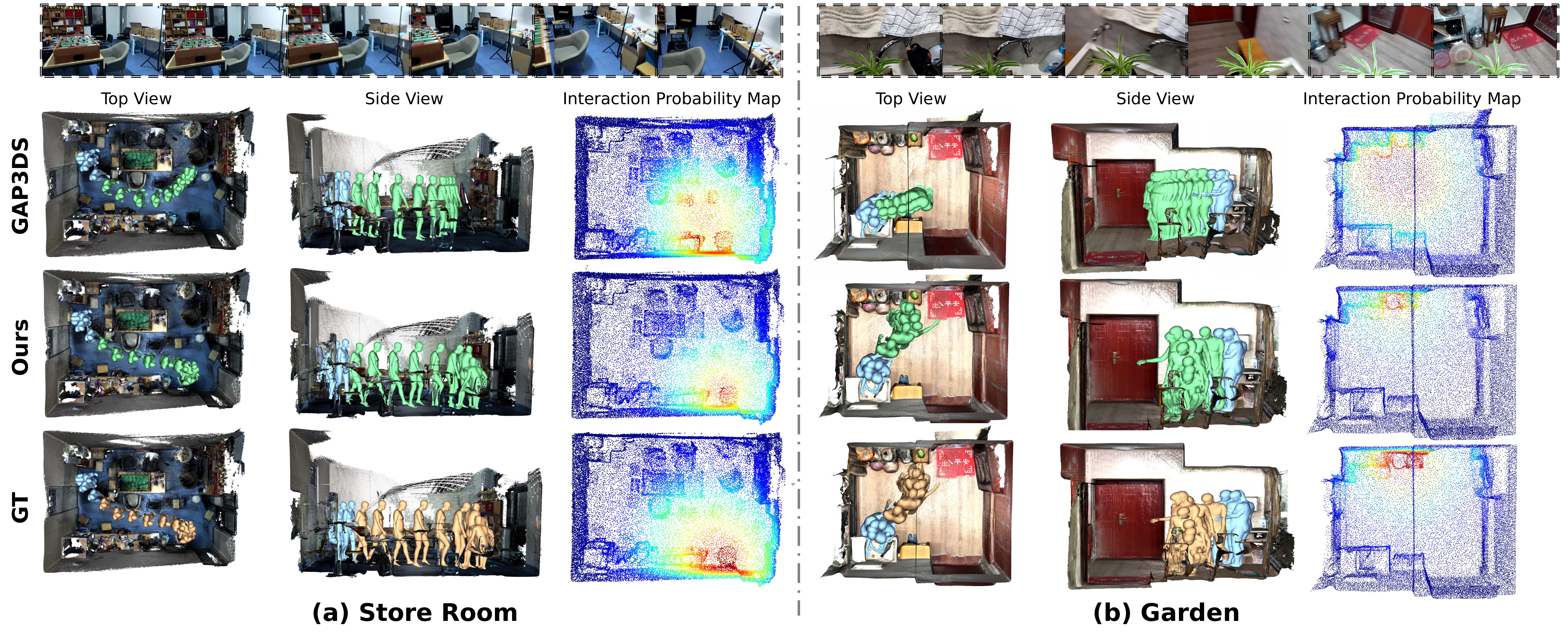}
    \vspace{-14pt}
    \caption{\textbf{Visualization Comparison.} The blue, green, and tan human meshes represent the historical, predicted, and ground-truth motion sequences, respectively. Zoom in for better visualization.}
  \label{fig:cpvis}
\end{figure}

As shown in Figure \ref{fig:cpvis}, we compare our EgoHMP with the state-of-the-art gaze-based method, GAP3DS \citep{yu2025vision}, across diverse scenarios. The rightmost columns for each scene visualize the predicted interaction probability maps. Due to the ambiguity and drift of human gaze signals, GAP3DS is easily misled by task-irrelevant visual regions, resulting in scattered and mislocalized interaction probability maps. Consequently, as observed in the top views and side views, this erroneous intent estimation causes GAP3DS to generate incorrect motion sequences. In contrast, EgoHMP accurately infers the user's interaction intent by exploring the egocentric visual context coupled with head motion cues. Guided by the precise interaction probability map, EgoHMP successfully generates context-consistent and physically plausible motion sequences.
\section{Conclusion and Discussion}
\label{Conclusion}
In this paper, we propose EgoHMP, a novel framework for precise human motion prediction (HMP) in 3D scenes. Recognizing the ambiguity, drift, and hardware dependency of gaze signals, we introduce a paradigm shift that uses egocentric observations and head motions as a more robust proxy for human interaction intent. To achieve this, EgoHMP bridges the cross-modal semantic gap and employs a modality-aware modulation mechanism to adaptively distill reliable interaction contexts. Crucially, to bridge scene perception and action execution, our framework simulates human "cognitive envisioning" by simultaneously predicting scene interaction and human contact regions. Guided by this dual-intent representation, our contact-aware motion decoder decouples global trajectory formulation from local pose synthesis. By utilizing a global trajectory as a self-prompt and progressively refining poses via a multi-stage denoiser, EgoHMP effectively mitigates geometric artifacts. Extensive experiments demonstrate that EgoHMP significantly outperforms existing methods, synthesizing 3D motions that are both semantically coherent and physically plausible. 

\textbf{Limitations and Future Work.} 
Currently, while egocentric views and head motions offer more robust intent modeling compared to gaze signals, extreme real-world cases (where both visual observations and head motions exhibit highly deceptive cues) will mislead the model's intent inference. Furthermore, EgoHMP predicts motion of a fixed sequence length. When the actual sequence length varies, it tends to truncate ongoing actions or generate stationary poses. Future work will explore long-term temporal causal reasoning to disambiguate short-term deceptive cues, as well as variable-length motion prediction paradigms to dynamically adapt to varying sequence length.

\section*{Acknowledgments}
This work was supported by the National Natural Science Foundation of China under Grants 62672066, 62272417, and 62672037. The authors declare no competing interests.
\newpage
\bibliographystyle{unsrtnat} 
\bibliography{references}    

\newpage
\appendix
\section*{Appendix}
\section{Architecture and Training Details}
\label{sec:architecture_details}
All training detailed in this section were conducted on a single NVIDIA RTX 3090 GPU. Notably, within the EgoHMP framework, only the Vision-Motion Alignment Encoder requires pre-training, whereas the Egocentric Intent Predictor and Contact-Aware Motion Decoder are optimized end-to-end. The necessity for this pre-training phase stems from the profound semantic and dimensional gap between 2D egocentric visual observations and 3D head motion kinematics. Jointly training such heterogeneous modalities from scratch alongside a complex downstream generative task can easily lead to representation collapse or suboptimal convergence. We conduct an experiment to validate this point. Please refer to Sec. \ref{Further Analysis}

\subsection{Pre-trained Vision-Motion Alignment Encoder}
\textbf{Vision Encoder.} The vision encoder takes egocentric video clips uniformly sampled at 2 fps. Each frame is resized to $224 \times 224$ and processed by a frozen pre-trained DINOv2 (ViT-L/14) \citep{oquab2023dinov2} model to extract frame-level features with a dimension of $D = 1024$. To model temporal correlations, we inject a learnable temporal positional embedding (initialized from a normal distribution $\mathcal{N}(0, 0.02)$) and process the sequence through a 4-layer Transformer encoder with 8 attention heads and a feed-forward network dimension of 2048. The global representation is obtained via average pooling along the temporal axis and then mapped to a latent space of $D_{proj} = 256$ using a projection head consisting of an MLP (Linear(1024, 512) $\rightarrow$ BatchNorm1d $\rightarrow$ GELU $\rightarrow$ Linear(512, 256)). 

\textbf{Motion Encoder.} For the motion encoder, the raw input is a 12-dimensional relative motion vector for each frame (comprising a 3D translation vector and a $3 \times 3$ flattened rotation matrix relative to the first frame). This input is first lifted to a 1024-dimensional space via a spatial MLP (Linear(12, 512) $\rightarrow$ LayerNorm $\rightarrow$ GELU $\rightarrow$ Linear(512, 1024)). Similar to the vision branch, temporal positional embeddings are added before feeding the sequence into a 1-layer lightweight Transformer encoder (4 attention heads, 2048 FFN dimension). The global motion representation is similarly average-pooled and mapped to $D_{proj} = 256$ via an identical projection head.

The vision-motion alignment encoder requires robust pre-training prior to the joint optimization of EgoHMP. To maximize alignment generalization and avoid data leakage, we propose a massive self-supervised data mining strategy. Specifically, we apply a dense sliding window with a stride of 10 frames across all training sequences to construct a large-scale paired vision-motion pre-training dataset. During this process, any sliding window that intersects with the test set sequences (or a 15-frame safety margin surrounding them) is strictly excluded.

\textbf{Training Details.} During pre-training, the model is optimized using the vision-motion alignment loss $\mathcal{L}_{A}$ defined in Eq. (2) of the main manuscript. The temperature hyper-parameters for the KL divergence and the InfoNCE loss (Cross-Entropy) are empirically set to $\tau_{kl} = 1.0$ and $\tau_{ce} = 0.7$, respectively. The balancing weight for the contrastive objective is set to $\lambda = 5.0$. The alignment encoder is pre-trained for 80 epochs using the Adam optimizer with a learning rate of $1\times 10^{-4}$ and a batch size of 64. We implement an offline caching mechanism for data pre-processing and feature extraction, which requires a one-time overhead of approximately 20 minutes. Once cached, the in-memory data loading reduces the time per epoch to merely 3 seconds, allowing the core pre-training phase to be completed in roughly 4 minutes.

\subsection{Egocentric Intent Predictor}
\label{sec:intent_predictor_details}

\textbf{Intent Transformer and Modality-Aware Modulation.} 
To adaptively extract reliable interaction contexts and suppress invalid cues, we apply the modality-aware modulation mechanism as formulated in the main manuscript. Specifically, we define learnable modality-specific modulation tokens, $\tau_{V}, \tau_{M} \in \mathbb{R}^{256}$, which scale the respective projected features via the Hadamard product ($\odot$). This multiplicative gating effectively functions as a gradient interception mechanism during backpropagation. Subsequently, learnable temporal positional embeddings ($E_{pos} \in \mathbb{R}^{1 \times 6 \times 256}$) are added to the modulated features to preserve temporal dynamics. The resulting visual and motion sequences are then concatenated along the temporal dimension. Finally, a learnable global intent token ($X_{intent} \in \mathbb{R}^{1 \times 1 \times 256}$) is prepended to the concatenated features, yielding a unified 13-token sequence. This sequence is processed by a 2-layer Transformer encoder (8 attention heads, 1024 FFN dimension) to establish dense cross-modal interactions, and the updated $X_{intent}$ token is extracted as the global intent representation.

\textbf{Scene Encoding and Interaction Probability Map Generation.} 
For the 3D scenes, we employ PointNet++ \citep{qi2017pointnet++} as our backbone to downsample and encode the raw 3D scene point clouds into point-wise geometric features, which are then mapped to the shared latent dimension ($D_{e} = 256$). To capture complex intra-scene topological relations, these point features are first refined by a 1-layer self-attention module (8 attention heads). Subsequently, we introduce a 3-layer cross-attention mechanism where the refined scene point features act as queries to attend to the 13-token intent representation (acting as keys and values). Ultimately, a prediction head consisting of a 1D convolution block (Conv1d(256, 64) $\rightarrow$ ReLU $\rightarrow$ Conv1d(64, 1)) projects the fused features into point-wise probability, yielding the final interaction probability map.

\textbf{Intent Pseudo-Label and Intent Loss.}
Since human cognitive intent is inherently unobservable and lacks explicit annotations in existing datasets, we formulate a pseudo-label generation strategy (Eq. (5) of the main manuscript.) to provide probabilistic supervision. To address potential concerns regarding the empirical fidelity and robustness of these hand-crafted heuristics, we ground our design in fundamental cognitive science and biomechanical principles. First, the pseudo-label leverages the future physical contact point as an absolute anchor, representing the ultimate physical manifestation of intent. Second, it incorporates head motion proximity. Extensive studies in motor control demonstrate that head orientation acts as a reliable kinematic precursor to hand and body interactions. Crucially, we introduce a temporal decay weight $(t/\Delta T)^2$ to modulate the head motion prior. This reflects the cognitive reality that as a human approaches an interaction moment, their head trajectory becomes increasingly deterministic and strongly correlated with the target, thereby minimizing ambiguity. Furthermore, our method is inherently robust to potential errors or ambiguities in these heuristics. Rather than employing rigid, one-hot hard labels, Eq. (5) yields a continuous, Gaussian-smoothed probability distribution map. Supervising the network via Kullback-Leibler (KL) divergence with these soft distributions acts as a form of label smoothing. It prevents the model from overfitting to isolated heuristic inaccuracies and encourages the network to learn the underlying multi-modal distribution of plausible intents. 

\subsection{Contact-Aware Motion Decoder}
\label{sec:motion_decoder_details}
\textbf{Trajectory Inference.} 
To decouple global locomotion from local pose articulation, we first predict a coarse future trajectory. The predicted interaction target (extracted via the argmax of the intent probability map) serves as the spatial goal. The historical 6-frame trajectory and the goal are individually embedded into a high-dimensional latent space ($D_{latent} = 768$) using linear projections. To maintain temporal continuity, we interpolate the intermediate frames, constructing a complete 16-frame sequence. This sequence undergoes successive cross-modal fusions: an 8-layer cross-attention module queries the global scene features, followed by another 8-layer cross-attention module that selectively queries non-obstacle geometric features. Finally, a 4-layer Transformer encoder (8 attention heads) refines these temporal features to output the predicted 6D global trajectories (3D translation and 3D axis-angle orientation). This predicted trajectory explicitly serves as a self-prompt for the subsequent diffusion process.

\textbf{Scene-Aware Conditioner.} 
To effectively ground the local pose generation within the 3D environment, we introduce a local scene conditioner. For each predicted frame, we transform the global point cloud into the local coordinate system of the current predicted root trajectory. We apply a cylindrical spatial mask---defined by a radius of 1.0 meter and a vertical range from -1.2 to 1.0 meters relative to the root---to filter out task-irrelevant geometries, retaining only the potential contact surfaces. To compress this local geometry into a compact representation, we initialize 4 learnable latent queries per frame. A Multi-Head Attention pooling layer (8 heads) aggregates the masked scene features into these queries, resulting in a highly efficient, contact-aware local context embedding sequence.

\textbf{Multi-Stage Denoiser.} 
The local pose generation is formulated as a conditional denoising diffusion probabilistic model. The denoising backbone is a 12-layer Transformer Decoder (12 attention heads, $D_{latent} = 768$, dropout $= 0.1$) that processes noisy latent poses conditioned on temporal embeddings, cross-modal features, and local scene contexts. To relieve physical artifacts, we design a multi-stage optimization strategy instead of a direct mapping. In the first stage, the network maps the latent features to predict a coarse 32D VPoser latent pose alongside a dense, vertex-level human contact probability map (10,475 vertices). In the second stage, we introduce an Explicit Kinematic Pooling mechanism. Utilizing official SMPL-X Linear Blend Skinning weights, the 10,475 vertex probabilities are rigorously pooled into 22 joint-level contact priors. These kinematic priors are then encoded to generate scale ($\gamma$) and shift ($\beta$) parameters for an Adaptive Layer Normalization (AdaLN), which modulates the latent representations to inject explicit contact awareness. Crucially, the AdaLN generator is zero-initialized to prevent noisy intent signals from disrupting the generative manifold during early training. In the final stage, the modulated latent features, concatenated with the global trajectory and the stop-gradient coarse pose, are processed by a residual refiner to predict a fine-grained latent displacement ($\Delta P$). The final refined pose is obtained by adding this zero-initialized residual to the coarse pose, successfully alleviating geometric artifacts such as floating and penetration while preserving the fundamental motion manifold.

\textbf{Human Contact Pseudo-Label and Contact Loss.}
To provide supervision for human contact learning, we derive pseudo-labels for human-scene contact, as explicit contact annotations are absent in datasets. Specifically, we decode the ground-truth motion sequence into explicit 3D human meshes using the SMPL-X model. For each frame $t$, we compute the shortest Euclidean distance between each of the $V = 10,475$ human mesh vertices, denoted as $v_{t,i}$, and the 3D scene point cloud $S$. We then formulate the contact pseudo-label $\mathcal{A}_{t,i}$ by mapping this distance into a soft probability distribution via a Gaussian kernel:
\begin{equation}
\mathcal{A}_{t,i} = \exp\left( -\frac{\min_{p \in S} \| v_{t,i} - p \|_2^2}{2\sigma^2} \right).
\end{equation}
Crucially, employing a Gaussian kernel rather than binary threshold simulates the physical nature of real-world human-scene interactions. For instance, when a human grasps a cup, the entire hand does not conform perfectly and uniformly to the surface; rather, the interaction manifests as focal contact points surrounded by a continuous spatial decay. By setting the standard deviation $\sigma$ to $0.02$ meters, our formulation accurately captures this proximity-based interaction pattern. Finally, after applying softmax normalization to ensure valid probability distributions, the contact loss $\mathcal{L}_{con}$ is defined as the KL divergence between the predicted contact map $\hat{\mathcal{A}}$ and the pseudo-label $\mathcal{A}$:
\begin{equation}
\mathcal{L}_{con} = \frac{1}{\Delta T} \sum_{t=T+1}^{T+\Delta T} \mathrm{KL}(\mathrm{softmax}(\mathcal{A}_t) \,||\, \mathrm{softmax}(\hat{\mathcal{A}}_t)).
\end{equation}
Through the contact region learning, EgoHMP effectively captures the continuous spatial relationship between the human body and the 3D environment, which is paramount for alleviating geometric artifacts such as floating and penetration.

\textbf{Motion Decoder.} 
To ensure physical plausibility and alleviate foot-sliding artifacts, the predicted 32D VPoser latent codes and the predicted 6D trajectories are decoded into 23 3D joint coordinates (69D) via the pre-trained VPoser and SMPL-X \citep{pavlakos2019expressive} models. The 3D joints and the latent codes are concatenated to form a 101D mixed node representation. This sequence is then processed by a 9-stage Graph Convolutional Network (GCN). To ensure temporal smoothness, the GCN uses Discrete Cosine Transform (DCT) and Inverse DCT (IDCT). The refined joints are ultimately aligned with the predicted trajectory to prevent root drift.

\textbf{Training Details.} 
The Egocentric Intent Predictor and Contact-Aware Motion Decoder are optimized in an end-to-end joint training paradigm. We employ the AdamW optimizer with an initial learning rate of $1 \times 10^{-4}$, governed by a cosine annealing scheduler across 120 epochs. Each epoch takes approximately 45 seconds, with the total training process completing in roughly 1.5 hours.

\section{Additional Experiments}
\begin{table}[!t]
    \centering
    \caption{\textbf{Quantitative Results.} Evaluated on the GIMO \citep{Zheng2022}. We analyze the necessity of the pre-training for the vision-motion alignment encoder and the impact of different objectives within the alignment loss. E2E denotes training the entire framework end-to-end. $\mathcal{L}_A$ denotes the vision-motion alignment loss. $kl$ denotes the KL divergence loss in $\mathcal{L}_A$. $nce$ denotes the InfoNCE contrastive loss in $\mathcal{L}_A$. R@1, R@5, and R@10 denote the cross-modal retrieval recall metrics (\%). $\mathcal{M} \rightarrow \mathcal{V}$ denotes motion-to-vision retrieval, and $\mathcal{V} \rightarrow \mathcal{M}$ denotes vision-to-motion retrieval.}
    \label{tab:ablation_pretrain}
    \resizebox{\textwidth}{!}{
    \begin{tabular}{l ccc ccc cccc ccc c} 
        \toprule
        \multirow{2}{*}{\textbf{Strategy}} 
        & \multicolumn{3}{c}{$\mathcal{M} \rightarrow \mathcal{V} \uparrow$}
        & \multicolumn{3}{c}{$\mathcal{V} \rightarrow \mathcal{M} \uparrow$}
        & \multicolumn{4}{c}{Motion Metrics $\downarrow$} 
        & \multicolumn{3}{c}{Intent Metrics $\downarrow$}
        & \multicolumn{1}{c}{Phys. $\downarrow$} \\
        \cmidrule(lr){2-4} \cmidrule(lr){5-7} \cmidrule(lr){8-11} \cmidrule(lr){12-14} \cmidrule(lr){15-15}
        & R@1 & R@5 & R@10 & R@1 & R@5 & R@10 & T-P & T-E & M-P & M-E & @1 & @50 & @100 & Pen. \\
        \midrule
        (a) E2E                            & 0.00 & 0.06 & 0.12 & 0.03 & 0.09 & 0.11 & 588 & 672 & 163.2 & 200.4 & 312.5 & 305.7 & 298.4 & 3.15 \\  
        (b) Pre-train w/o $kl$             & 57.53 & 90.90 & 95.53 & 55.95 & 92.23 & 96.77 & 580 & 647 & 151.6 & 189.1 & 295.4 & 288.6 & 282.5 & 2.42 \\  
        (c) Pre-train w/o $nce$            & 0.00 & 0.04 & 0.11 & 0.01 & 0.10 & 0.13 & 490 & 558 & 119.4 & 144.2 & 188.5 & 180.2 & 174.0 & 0.89 \\  
        \midrule
        \rowcolor{gray!15} 
        \textbf{(d) Full Pre-training}     & \textbf{66.37} & \textbf{94.75} & \textbf{97.84} & \textbf{63.25} & \textbf{95.44} & \textbf{98.15} & \textbf{481} &\textbf{545} &\textbf{114.9} &\textbf{138.0} &\textbf{161.3} &\textbf{153.2} &\textbf{148.6} &\textbf{0.81} \\
        \bottomrule
    \end{tabular}
    }
\end{table}

\subsection{Further Analysis of Vision-Motion Alignment Encoder}
\label{Further Analysis}

\textbf{Pre-training.} 
As shown in Table \ref{tab:ablation_pretrain} (a), when directly training the entire framework end-to-end (E2E) without pre-training phase, the model's performance drops drastically, highlighted by a near-zero cross-modal retrieval recall. This degradation fundamentally stems from the severe risk of dimensional collapse during joint optimization. Specifically, the continuous fine-tuning of the visual backbone causes the discrepancies among visual features to fluctuate violently. Consequently, to minimize the penalty, the alignment loss forces the visual disparities to zero, which in turn drives the motion disparities to zero as well. Furthermore, 2D visual observations and 3D head kinematics inherently reside in highly heterogeneous semantic spaces. Forcing joint optimization without prior feature alignment leads to severe multi-modal feature collision, thereby destroying the discriminability of the feature space and providing ambiguous, noisy cues to the downstream generation modules.

\textbf{Vision-Motion Alignment Loss.} 
Table \ref{tab:ablation_pretrain} (b) and (c) demonstrates the indispensable and complementary roles of our composite objectives. Discarding the InfoNCE contrastive loss (w/o $nce$) leads to a failure in cross-modal retrieval. This confirms that InfoNCE provides the primary driving force for establishing instance-level alignment between visual and kinematic signals. Conversely, relying solely on InfoNCE without the KL divergence loss (w/o $kl$) yields high retrieval metrics but severely degrades downstream motion generation  and physical plausibility. Without the distribution regularization provided by KL divergence, the latent feature manifold becomes sharply overfitted to the retrieval objective, failing to provide the smooth and continuous prior essential for the downstream motion generator. Ultimately, the full composite loss achieves a profound synergy: KL divergence smooths the latent space to support robust generation, which in turn alleviates feature overfitting.
\begin{table}[!t]
    \centering
    \setlength{\tabcolsep}{4pt} 
    \caption{\textbf{Quantitative Results.} Trajectory deviation (Traj) and pose error (MPJPE) across seven  3D scenes on the GIMO \citep{Zheng2022}. Time horizons ($0.5$, $2.0$, $5.0$) are reported in (s). Best results are in \textbf{bold}.}
    \label{tab:quantitative_comparison_scenarios_single}
    \resizebox{\textwidth}{!}{
    \begin{tabular}{l l ccc ccc ccc ccc ccc ccc ccc}
        \toprule
        \multirow{2}{*}{\textbf{Method}} & \multirow{2}{*}{\textbf{Metric}} 
        & \multicolumn{3}{c}{\textbf{Bedroom}} & \multicolumn{3}{c}{\textbf{Classroom}} & \multicolumn{3}{c}{\textbf{Garden}} & \multicolumn{3}{c}{\textbf{Kitchen}} & \multicolumn{3}{c}{\textbf{Lab}} & \multicolumn{3}{c}{\textbf{Livingroom}} & \multicolumn{3}{c}{\textbf{Seminarroom}} \\  
        \cmidrule(lr){3-5} \cmidrule(lr){6-8} \cmidrule(lr){9-11} \cmidrule(lr){12-14} \cmidrule(lr){15-17} \cmidrule(lr){18-20} \cmidrule(lr){21-23}
        & & 0.5 & 2.0 & 5.0 & 0.5 & 2.0 & 5.0 & 0.5 & 2.0 & 5.0 & 0.5 & 2.0 & 5.0 & 0.5 & 2.0 & 5.0 & 0.5 & 2.0 & 5.0 & 0.5 & 2.0 & 5.0 \\  
        \midrule
        \multirow{2}{*}{BiFu \citep{Zheng2022}} 
        & Traj  & 215 & 495 & 735 & 312 & 425 & 605 & 202 & 500 & 578 & 257 & 440 & 515 & 256 & 515 & 595 & 317 & 580 & 702 & 394 & 622 & 558 \\
        & MPJPE & \textbf{62.0} & 106.5 & 162.6 & 84.1 & 136.0 & 151.3 & 83.2 & 193.0 & 223.5 & 91.5 & 100.7 & 118.1 & \textbf{49.5} & 106.2 & 203.2 & \textbf{89.4} & 119.8 & 210.2 & 100.7 & 131.0 & 161.5 \\
        \cmidrule(lr){1-23}
        \multirow{2}{*}{SIF3D \citep{Lou2024}} 
        & Traj  & 171 & 415 & 485 & \textbf{145} & 405 & 482 & 186 & 425 & 565 & 165 & 445 & 395 & 231 & 505 & 645 & 241 & 565 & 698 & 205 & 415 & 542 \\
        & MPJPE & 77.5 & 98.4 & 162.6 & 93.3 & 138.0 & 141.3 & \textbf{68.5} & 176.2 & 175.1 & 93.1 & 97.4 & 101.5 & 61.4 & 88.1 & 191.9 & 102.2 & \textbf{115.3} & 166.2 & 94.7 & 131.1 & 158.6 \\
        \cmidrule(lr){1-23}
        \multirow{2}{*}{GAP3DS \citep{yu2025vision}} 
        & Traj  & 168.5 & \textbf{350} & 415 & 179 & 391 & 413 & 183 & \textbf{422} & 472 & 163 & \textbf{377} & 368 & 235 & 465 & 560 & 173 & \textbf{515} & 576 & 174 & 375 & 531 \\
        & MPJPE & 88.4 & \textbf{93.8} & 152.4 & 93.2 & 117.3 & 125.2 & 84.2 & \textbf{164.3} & 169.0 & 84.4 & \textbf{85.2} & 88.6 & 53.6 & 87.9 & 140.0 & 96.2 & 119.0 & 155.8 & 96.8 & 119.4 & 153.1 \\
        
        \midrule
        \multirow{2}{*}{Auxformer \citep{Xu2023}} 
        & Traj  & 171 & 519 & 1022 & 295 & 424 & 465 & 211 & 592 & 851 & 162 & 510 & 552 & 232 & 675 & 977 & 140 & 588 & 905 & 252 & 569 & 785 \\
        & MPJPE & 106.1 & 168.3 & 228.5 & 128.0 & 169.9 & 143.5 & 97.5 & 217.6 & 203.1 & 119.1 & 108.5 & 112.6 & 105.3 & 132.1 & 205.2 & 146.3 & 127.5 & 230.1 & 122.9 & 151.8 & 181.5 \\
        \cmidrule(lr){1-23}
        \multirow{2}{*}{MDP \citep{xing2024scene}} 
        & Traj  & \textbf{154} & 515 & 592 & 165 & 415 & 506 & 181 & 455 & 565 & 155 & 425 & 390 & 244 & 575 & 602 & 230 & 575 & 738 & \textbf{158} & 481 & 561 \\
        & MPJPE & 107.0 & 169.0 & 247.6 & 89.4 & 155.8 & 144.5 & 94.3 & 177.2 & 185.3 & 85.9 & 100.0 & 110.5 & 74.5 & \textbf{87.4} & 203.9 & 95.7 & 124.1 & 185.9 & 96.7 & 133.0 & 166.1 \\
        \midrule
        \rowcolor{gray!15} 
        & Traj  & 163 & 351 & \textbf{403} & 157 & \textbf{278} & \textbf{352} & \textbf{170} & 435 & \textbf{461} & \textbf{155} & 395 & \textbf{359} & \textbf{221} & \textbf{430} & \textbf{540} & \textbf{161} & 528 & \textbf{554} & 166 & \textbf{352} & \textbf{492} \\
        \rowcolor{gray!15} 
        \multirow{-2}{*}{\textbf{EgoHMP (Full)}} 
        & MPJPE & 86.0 & 94.0 & \textbf{146.1} & \textbf{82.5} & \textbf{95.2} & \textbf{119.9} & 80.0 & 166.2 & \textbf{168.4} & \textbf{81.4} & 85.9 & \textbf{88.1} & 53.5 & \textbf{87.4} & \textbf{132.7} & 89.6 & 126.9 & \textbf{144.7} & \textbf{91.1} & \textbf{108.0} & \textbf{136.5} \\
        \bottomrule
    \end{tabular}
    }
\end{table}
\subsection{Analysis of Prediction Horizons}
We evaluate the performance of EgoHMP against several baselines across various prediction horizons. As illustrated in Table \ref{tab:quantitative_comparison_scenarios_single}, while most methods maintain reasonable accuracy in short-term forecasting due to physical inertia, baseline performance deteriorates significantly as the temporal horizon expands, leading to substantial error accumulation. In contrast, EgoHMP achieves superior performance across all metrics in both mid- and long-term horizons. This improvement is attributed to our construction of precise interaction probability maps, which effectively mitigates the error accumulation that plagues other methods due to inaccurate initial intent localization.
 
\subsection{Discussion on Stochastic and Deterministic Paradigms}
In this section, we provide a comprehensive discussion of stochastic and deterministic human motion prediction (HMP) methods to contextualize EgoHMP within existing paradigms.

\textbf{Stochastic Scene-Agnostic Paradigms.} 
Methods such as DLow~\citep{yuan2020dlow}, SmoothDMP~\citep{mao2021generating}, BeLFusion~\citep{barquero2023belfusion}, and STAB-ACB~\citep{tang2025stochastic} prioritize modeling the multimodal distribution of future poses. By leveraging generative backbones (e.g., VAEs, Diffusion, or Flow-based models), they synthesize diverse motion hypotheses based primarily on historical sequences. While achieving high diversity scores, these methods often function as ``blind'' generators. Due to the absence of explicit grounding in 3D scene geometry, the generated diversity frequently manifests as physically implausible random drifts or collisions. In essence, they prioritize kinematic variety at the expense of interaction fidelity.
\begin{figure}
\centering
\includegraphics[width=\linewidth,scale=1]{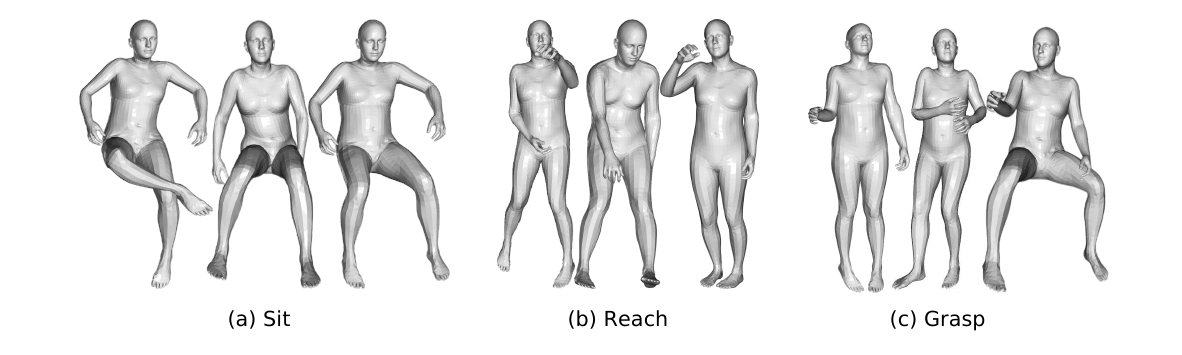}
    \vspace{-20pt}
    \caption{\textbf{Qualitative Visualization.} Given specific interaction intents, EgoHMP synthesizes diverse poses, confirming effective exploration of the valid interaction subspace without mean pose collapse.}
  \label{fig:multipose}
\end{figure}

\textbf{Deterministic and Regression-based Scene-Aware Paradigms.} 
Conversely, methods like Auxformer~\citep{Xu2023}, MDP~\citep{xing2024scene}, BiFu~\citep{Zheng2022}, and SIF3D~\citep{Lou2024} integrate scene geometry to guide prediction through regression-based decoders. While these methods achieve better stability, they suffer from mean pose problem. By mapping complex poses to a single deterministic sequence, they tend to converge to the most generic interaction, failing to capture the rich variations of human behavior. 

\textbf{Reconciling Stochastic and Deterministic Paradigms.} 
To effectively reconcile the trade-off between stochastic exploration and deterministic guidance, recent advanced methods like GAP3DS \citep{yu2025vision} and our EgoHMP adopt a two-stage, intent-guided generation paradigm. Unlike  stochastic methods, this paradigm first establishes a deterministic global trajectory derived from predicted intent, effectively pruning task-irrelevant hypotheses. Building upon this, the diffusion-based decoder functions as a stochastic engine to synthesize kinematically diverse poses compliant with the intent. This paradigm allows model to explore effective interaction subspaces without collapsing into a singular mean pose. 

\textbf{Dataset: CIRCLE} \citep{araujo2023circle} comprises 10 hours of high-fidelity full-body motion sequences captured from 5 subjects across nine apartment scenes. Leveraging a 12-camera Vicon system at 120 FPS integrated with the AI Habitat VR environment, it ensures precise synchronization between human motion and scene geometry. The dataset utilizes integrated apartment meshes, treating each room as a distinct scene, and contains over 7,000 sequences covering 128 diverse tasks.

\textbf{Baselines.} We benchmark EgoHMP against five methods categorized into two paradigms. To represent purely stochastic scene-agnostic methods, we select DLow~\citep{yuan2020dlow} (Flow-based), SmoothDMP~\citep{mao2021generating} (VAE-based), BeLFusion~\citep{barquero2023belfusion} (Diffusion-based), and STAB-ACB~\citep{tang2025stochastic} (Memory-based), which encompass the state-of-the-art in diverse motion generation. To represent the paradigm that reconciles stochastic exploration with deterministic guidance, we select the GAP3DS~\citep{yu2025vision}. 

\textbf{Metrics.} Following DLow~\citep{yuan2020dlow}, we quantify the prediction diversity through the average pairwise distance (APD), computed as the mean L2 distance between the generated samples. To evaluate reconstruction accuracy, we employ average displacement error (ADE) and final displacement error (FDE) to measure whole-sequence and endpoint quality, respectively. Furthermore, we report MMADE and MMFDE, to assess the fidelity of the best-matched hypothesis in diverse scenarios.

\begin{wraptable}{r}{8cm} 
    \centering
    \vspace{-18pt}
    \caption{\textbf{Quantitative Results.} $\uparrow$: higher is better, $\downarrow$: lower is better. The best results are highlighted in \textbf{bold}.}
    \label{tab:circle_compact}
    \resizebox{\linewidth}{!}{
        \begin{tabular}{lccccc}
            \toprule
            Method & APD $\uparrow$ & ADE $\downarrow$ & FDE $\downarrow$ & MMADE $\downarrow$ & MMFDE $\downarrow$ \\
            \midrule
            DLow \citep{yuan2020dlow}& 49.37 & 13.49 & 14.95 & 11.70 & 14.46 \\
            SmoothDMP \citep{mao2021generating}& \textbf{57.38} & 11.31 & 13.57 & 9.75 & 12.04 \\
            BeLFusion \citep{barquero2023belfusion}& 39.46 & 9.56 & 10.04 & 7.91 & 9.30 \\
            STAB-ACB \citep{tang2025stochastic}& 52.11 & 11.49 & 13.95 & 11.06 & 12.14 \\
            GAP3DS \citep{yu2025vision}& 42.47 & 5.37 & 6.51 & 4.85 & 5.62 \\
            \midrule
            \rowcolor{gray!15} 
            \textbf{EgoHMP (Full)} & 44.10 & \textbf{4.98} & \textbf{6.17} & \textbf{4.12} & \textbf{5.08} \\
            \bottomrule
        \end{tabular}
    }
    \vspace{-12pt}
\end{wraptable}
\textbf{Results and Analysis.} As shown in Table \ref{tab:circle_compact}, purely stochastic and scene-agnostic methods (e.g., SmoothDMP, STAB-ACB) achieve high diversity scores by unconstrainedly exploring the latent motion space. However, they suffer from severe interaction infidelity. Despite advancements in diffusion (BeLFusion) and memory-based priors that enhance kinematic consistency, their elevated deterministic (ADE/FDE) and multimodal errors (MMADE/MMFDE) confirm that internal motion priors are insufficient without explicit environmental grounding. In essence, they function as ``blind'' generators, producing physically implausible random drifts to inflate diversity. In contrast, the intent-guided paradigms (GAP3DS and EgoHMP) drastically reduce displacement errors. Notably, EgoHMP achieves state-of-the-art performance across all fidelity metrics. EgoHMP uses egocentric visual and motion cues to construct a precise intent-driven trajectory. This robust trajectory prevents the model from diverging into invalid physical spaces. Consequently, EgoHMP maintains competitive local pose diversity while strictly confining the generative manifold to the true intent distribution. As demonstrated in Figure \ref{fig:multipose}, given specific interaction intents (e.g., sitting, reaching, or grasping), EgoHMP synthesizes diverse poses, confirming effective exploration of the valid interaction subspace without mean pose collapse.

\subsection{Analysis of Vision Backbone and Computational Efficiency}
\begin{table}[t]
    \centering
    \caption{\textbf{Quantitative Results.} VRAM denotes the GPU memory consumption (in MB), and Speed represents the inference speed (in samples/s). The best results are highlighted in \textbf{bold}.}
    \label{tab:ablation_efficiency}
    \resizebox{\textwidth}{!}{
        \begin{tabular}{l | ccc>{\columncolor{gray!10}}cc | cc>{\columncolor{gray!10}}cc}
            \toprule
            \multirow{2}{*}{\textbf{Metric}} & \multicolumn{5}{c|}{\textbf{Vision Backbone (with 16384 points)}} & \multicolumn{4}{c}{\textbf{Scene Points (with DINOv2-L)}} \\
            \cmidrule(lr){2-6} \cmidrule(lr){7-10}
            & ResNet-50 \citep{he2016deep}& CLIP \citep{radford2021learning}  & DINOv2-B \citep{oquab2023dinov2}& DINOv2-L \citep{oquab2023dinov2} & DINOv3-L \citep{simeoni2025dinov3} & 4096 & 8192 & 16384 & 32768 \\
            \midrule
            \textbf{T-P} ($\downarrow$) & 572  & 534 & 512 & 481 & \textbf{479} & 538 & 511 & 481 & \textbf{480} \\
            \textbf{M-P} ($\downarrow$)  & 138.6 & 133.2 & 124.5 & 114.9 & \textbf{113.2} & 129.4 & 121.5 & 114.9 & \textbf{114.3} \\
            \textbf{VRAM} ($\downarrow$) & \textbf{12100}  & 12650 & 12680 & 14610 & 14850 & \textbf{7840} & 11120 & 14610 & 18067 \\
            \textbf{Speed} ($\uparrow$) & \textbf{17.5}  & 16.8 & 16.7 & 15.1 & 14.8 & \textbf{27.4} & 20.8 & 15.1 & 8.9 \\
            \bottomrule
        \end{tabular}
    }
\end{table}
\textbf{Vision Backbone.} 
To investigate the impact of visual feature extractors on EgoHMP, we evaluate various backbones at a fixed scale of 16,384 scene points. As shown in the left section of Table \ref{tab:ablation_efficiency}, standard supervised architectures like ResNet-50 \citep{he2016deep} struggle to capture the fine-grained semantic required for human-scene interactions, yielding the highest trajectory and pose errors. While vision-language models like CLIP \citep{radford2021learning} provide stronger high-level semantics, they lack dense geometric understanding. Notably, employing self-supervised foundation models (DINOv2 \citep{oquab2023dinov2}) significantly enhances predictive performance. Scaling from Base (DINOv2-B) to Large (DINOv2-L) further reduces the prediction error, albeit with a moderate increase in VRAM. Although the latest DINOv3-L \citep{simeoni2025dinov3} achieves the best prediction accuracy, its performance gain over DINOv2-L is marginal compared to the additional computational overhead. Therefore, to achieve the optimal trade-off between accuracy and efficiency, we adopt DINOv2-L as our vision backbone.

\textbf{Scene Points.} 
As shown in the right section of Table \ref{tab:ablation_efficiency}, we further explore the impact of the sampled scene point number on model accuracy and inference performance. As the number of points increases from 4,096 to 16,384, the model's perception of the 3D environment improves significantly, leading to a sharp decrease in prediction errors. However, this performance gain eventually saturates. Forcing the point count to 32,768 yields only a marginal improvement in fidelity, but introduces a severe computational penalty and inference speed drops drastically. Consequently, we employ 16,384 points to strike the optimal balance between generative accuracy and inference efficiency.
\begin{figure}
\centering
\includegraphics[width=\linewidth,scale=1]{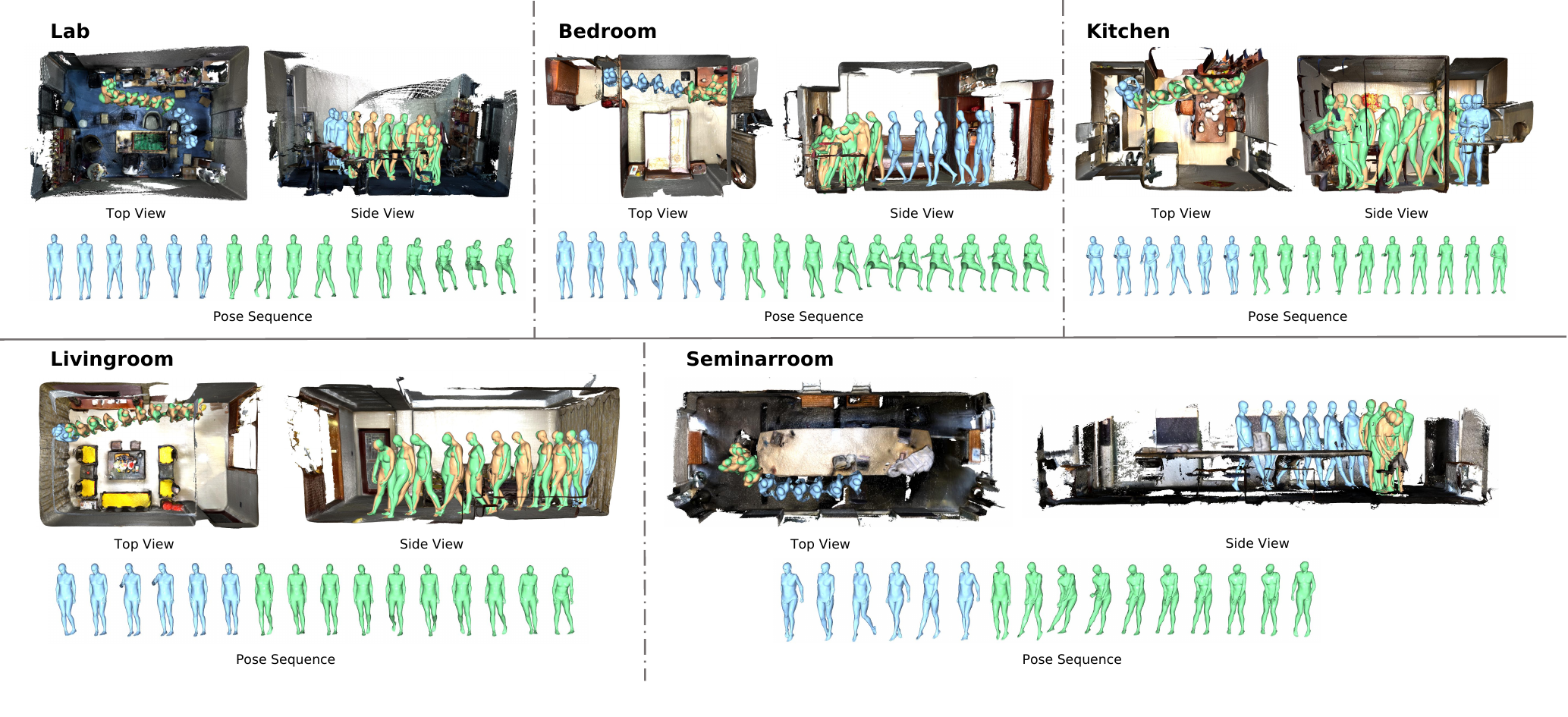}
    \vspace{-14pt}
    \caption{\textbf{Qualitative Visualization.} The blue, green, and tan human meshes represent the historical, predicted, and ground-truth motion sequences, respectively. Zoom in for better visualization.}
  \label{fig:app}
\end{figure}
\section{More Visualizations}
In this section, we provide more visualization results to demonstrate the capability of EgoHMP across various complex scenarios. Figure \ref{fig:app} illustrates the generated motion sequences in five scene types: lab, bedroom, kitchen, livingroom, and seminarroom.

As shown in the top view and side view, our model effectively perceives the 3D scene geometry and generates motions that strictly adhere to environmental constraints. In the kitchen scenario, the model successfully predicts a path through a narrow corridor between counters, avoiding collisions with the surrounding cabinets. 
Similarly, in the seminarroom, the generated character navigates alongside a long conference table, maintaining a natural distance from the furniture. The pose sequence strips at the bottom of each sub-figure demonstrate the kinematic quality of the predictions. The transition from the historical observation (blue meshes) to the predicted future (green meshes) is seamless, exhibiting no jitter or discontinuity. Whether the action involves straight walking  or complex turning maneuvers, the generated poses maintain contextual consistency and physical plausibility. 

\section{Potential Broader Impacts}
\label{PBI}

\textbf{Enhanced Human-Robot Collaboration.} EgoHMP can improve human-robot collaboration. By accurately predicting human motions, collaborative robots can proactively avoid collisions and mitigate unsafe interactions, thereby fostering safer environments for both humans and machines.

\textbf{Improved AR/VR and Immersive Experiences.} In augmented and virtual reality, this framework can drive more realistic and responsive interactions with virtual elements. By deeply understanding and predicting how a user might move and interact within a scene, AR/VR systems can offer highly immersive, natural experiences that seamlessly align with user expectations, ultimately enhancing user engagement.

\textbf{Advancements in Wearable Assistive Technologies.} For elderly individuals or people with physical disabilities, wearable assistive devices equipped with EgoHMP can anticipate daily needs or critical actions (e.g., reaching for an object or potential loss of balance). This capability enables timely physical or cognitive assistance, enhancing users' independence and overall quality of life.

\textbf{Applications in Autonomous Systems.} Integrating robust human motion prediction frameworks into broader autonomous systems (such as embodied AI and intelligent vehicles) can improve safety and traffic management. By proactively anticipating pedestrian or user movements, these autonomous agents can navigate complex, dynamic urban and indoor environments more safely and efficiently.

\newpage
\section*{NeurIPS Paper Checklist}

\begin{enumerate}

\item {\bf Claims}
    \item[] Question: Do the main claims made in the abstract and introduction accurately reflect the paper's contributions and scope?
    \item[] Answer: \answerYes{} 
    \item[] Justification: The main claims and contributions outlined in the Abstract and Introduction (Section \ref{Introduction}) are explicitly supported by the methodological formulation in Section \ref{Method} and the comprehensive experimental evaluations presented in Section \ref{Experiment}.

\item {\bf Limitations}
    \item[] Question: Does the paper discuss the limitations of the work performed by the authors?
    \item[] Answer: \answerYes{} 
    \item[] Justification: We analyze the limitation of the proposed EgoHMP in Section \ref{Conclusion}.

\item {\bf Theory assumptions and proofs}
    \item[] Question: For each theoretical result, does the paper provide the full set of assumptions and a complete (and correct) proof?
    \item[] Answer: \answerNA{} 
    \item[] Justification: This paper focuses on network design and evaluation for egocentric human motion prediction, and does not claim any formal theoretical results or mathematical proofs. The main results of this paper are experimental results.

    \item {\bf Experimental result reproducibility}
    \item[] Question: Does the paper fully disclose all the information needed to reproduce the main experimental results of the paper to the extent that it affects the main claims and/or conclusions of the paper (regardless of whether the code and data are provided or not)?
    \item[] Answer: \answerYes{} 
    \item[] Justification: All necessary information for reproduction, including network architecture, datasets, evaluation metrics, and training hyperparameters, is provided in Section \ref{Experiment} and Appendix \ref{sec:architecture_details}.

\item {\bf Open access to data and code}
    \item[] Question: Does the paper provide open access to the data and code, with sufficient instructions to faithfully reproduce the main experimental results, as described in supplemental material?
    \item[] Answer: \answerYes{} 
    \item[] Justification: The supplemental material provides the code.

\item {\bf Experimental setting/details}
    \item[] Question: Does the paper specify all the training and test details (e.g., data splits, hyperparameters, how they were chosen, type of optimizer) necessary to understand the results?
    \item[] Answer: \answerYes{} 
    \item[] Justification: All necessary information is provided in Section \ref{Experiment} and Appendix \ref{sec:architecture_details}.

\item {\bf Experiment statistical significance}
    \item[] Question: Does the paper report error bars suitably and correctly defined or other appropriate information about the statistical significance of the experiments?
    \item[] Answer: \answerNo{} 
    \item[] Justification: The performance margins between our method and the baselines are enough to demonstrate the effectiveness of our proposed approach.

\item {\bf Experiments compute resources}
    \item[] Question: For each experiment, does the paper provide sufficient information on the computer resources (type of compute workers, memory, time of execution) needed to reproduce the experiments?
    \item[] Answer: \answerYes{} 
    \item[] Justification: Please see Appendix \ref{sec:architecture_details} and Table \ref{tab:ablation_efficiency}.
    
\item {\bf Code of ethics}
    \item[] Question: Does the research conducted in the paper conform, in every respect, with the NeurIPS Code of Ethics \url{https://neurips.cc/public/EthicsGuidelines}?
    \item[] Answer: \answerYes{} 
    \item[] Justification: We have ensured that our research conforms to it.

\item {\bf Broader impacts}
    \item[] Question: Does the paper discuss both potential positive societal impacts and negative societal impacts of the work performed?
    \item[] Answer: \answerYes{} 
    \item[] Justification: Please refer to Appendix \ref{PBI} for a detailed discussion on the potential broader impacts of EgoHMP.
    
\item {\bf Safeguards}
    \item[] Question: Does the paper describe safeguards that have been put in place for responsible release of data or models that have a high risk for misuse (e.g., pre-trained language models, image generators, or scraped datasets)?
    \item[] Answer: \answerNA{} 
    \item[] Justification: This paper does not present any of the aforementioned risks.

\item {\bf Licenses for existing assets}
    \item[] Question: Are the creators or original owners of assets (e.g., code, data, models), used in the paper, properly credited and are the license and terms of use explicitly mentioned and properly respected?
    \item[] Answer: \answerYes{} 
    \item[] Justification: Please refer to the Reference.

\item {\bf New assets}
    \item[] Question: Are new assets introduced in the paper well documented and is the documentation provided alongside the assets?
    \item[] Answer: \answerNA{} 
    \item[] Justification: This paper does not introduce new assets.

\item {\bf Crowdsourcing and research with human subjects}
    \item[] Question: For crowdsourcing experiments and research with human subjects, does the paper include the full text of instructions given to participants and screenshots, if applicable, as well as details about compensation (if any)? 
    \item[] Answer: \answerNA{} 
    \item[] Justification: There is no crowdsourcing experiments and research with human subjects.

\item {\bf Institutional review board (IRB) approvals or equivalent for research with human subjects}
    \item[] Question: Does the paper describe potential risks incurred by study participants, whether such risks were disclosed to the subjects, and whether Institutional Review Board (IRB) approvals (or an equivalent approval/review based on the requirements of your country or institution) were obtained?
    \item[] Answer: \answerNA{} 
    \item[] Justification: There is no research with human subjects.

\item {\bf Declaration of LLM usage}
    \item[] Question: Does the paper describe the usage of LLMs if it is an important, original, or non-standard component of the core methods in this research? Note that if the LLM is used only for writing, editing, or formatting purposes and does \emph{not} impact the core methodology, scientific rigor, or originality of the research, declaration is not required.
    \item[] Answer: \answerNA{} 
    \item[] Justification: The core method development in this paper does not involve LLMs as any important, original, or non-standard components.

\end{enumerate}

\end{document}